\documentclass[letterpaper]{article} 
\usepackage{aaai2026}  
\usepackage{times}  
\usepackage{helvet}  
\usepackage{courier}  
\usepackage[hyphens]{url}  
\usepackage{graphicx} 
\usepackage{booktabs}
\usepackage{multirow} 
\usepackage{adjustbox}
\usepackage{natbib}  
\usepackage{caption} 
\usepackage{algorithm}
\usepackage{algorithmic}
\usepackage{amssymb}
\usepackage{amsmath}
\usepackage{bm}
\usepackage{newfloat}
\usepackage{listings}
\DeclareCaptionStyle{ruled}{labelfont=normalfont,labelsep=colon,strut=off} 
\floatstyle{ruled}
\newfloat{listing}{tb}{lst}{}
\floatname{listing}{Listing}
\title{Libra-MIL: Multimodal Prototypes Stereoscopic Infused with Task-specific Language Priors for Few-shot Whole Slide Image Classification}
\author{
    Zhenfeng Zhuang\textsuperscript{\rm 1}\equalcontrib,
    Fangyu Zhou\textsuperscript{\rm 1}\equalcontrib,
    Liansheng Wang\textsuperscript{\rm 1,2}\thanks{Corresponding author.}
}
\affiliations{
    \textsuperscript{\rm 1} Department of Computer Science at School of Informatics, Xiamen University, Xiamen, China.\\
    \textsuperscript{\rm 2} National Institute for Data Science in Health and Medicine, Xiamen University, Xiamen, China. \\
    \{zhuangzhenfeng, zhoufangyu\}@stu.xmu.edu.cn, lswang@xmu.edu.cn
}

\usepackage{bibentry}

\begin{document}

\maketitle

\begin{figure*}[ht!]
\includegraphics[width=0.98\textwidth]{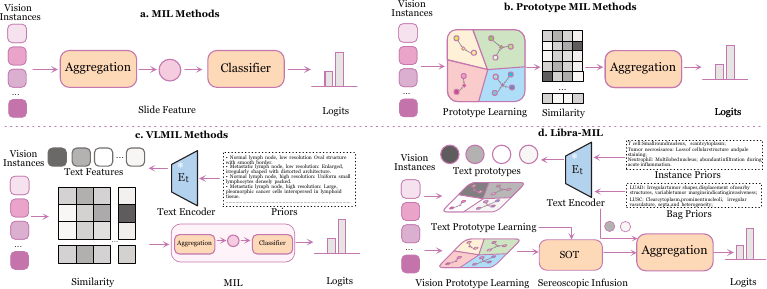} 
\centering
\caption{Brief comparison with related MIL methods. a) Basic MIL simply performs aggregation and sorting operations. b) Prototype MIL learns class-aware logits via instance similarity directly. c) VLMIL fused LLM Priors with the text-guided similarity. d) Libra-MIL uses Multimodal prototype learning with Stereoscopic Optimal Transport (SOT) on similarities.}
\label{intro}
\end{figure*}

\begin{abstract}
While Large Language Models (LLMs) are emerging as a promising direction in computational pathology, the substantial computational cost of giga-pixel Whole Slide Images (WSIs) necessitates the use of Multi-Instance Learning (MIL) to enable effective modeling. A key challenge is that pathological tasks typically provide only bag-level labels, while instance-level descriptions generated by LLMs often suffer from bias due to a lack of fine-grained medical knowledge. To address this, we propose that constructing task-specific pathological entity prototypes is crucial for learning generalizable features and enhancing model interpretability. Furthermore, existing vision-language MIL methods often employ unidirectional guidance, limiting cross-modal synergy. In this paper, we introduce a novel approach, Multimodal Prototype-based Multi-Instance Learning, that promotes bidirectional interaction through a balanced information compression scheme. Specifically, we leverage a frozen LLM to generate task-specific pathological entity descriptions, which are learned as text prototypes. Concurrently, the vision branch learns instance-level prototypes to mitigate the model's reliance on redundant data. For the fusion stage, we employ the Stereoscopic Optimal Transport (SOT) algorithm, which is based on a similarity metric, thereby facilitating broader semantic alignment in a higher-dimensional space. We conduct few-shot classification and explainability experiments on three distinct cancer datasets, and the results demonstrate the superior generalization capabilities of our proposed method. 
\end{abstract}

\begin{links}
    \link{Code, Appendix}{https://github.com/zfy07/Libra-MIL}
\end{links}

\section{Introduction}

AI-based histopathology is a promising direction for assisting diagnostics, especially in cancer diagnosis and grading \cite{omar2024chatgpt}. H\&E-stained WSIs, despite compression, have extremely high resolution (e.g., 40$\times$ magnification, 0.25$\mu$m/pixel, approximately 40,000 $\times$ 40,000 pixels) and large storage needs, challenging end-to-end training on fine-grained pixel data. Multiple instance learning (MIL), a weakly-supervised framework, addresses sparse instance-level labels (typically only WSI-level) and resource consumption in this area \cite{gadermayr2024multiple}.


In MIL paradigm, WSIs are treated as bags and their constituent patches as instances. Typically, pretrained feature extractors (e.g., UNI \cite{chen2024uni}, CONCH \cite{lu2024avisionlanguage}) are employed to embed these patches, and the resulting features are then aggregated to obtain bag-level embeddings for WSI-level prediction tasks. Although MIL-based approaches have achieved strong performance, their diagnostic capabilities have shown signs of saturation. To overcome this limitation, recent studies have explored the integration of multimodal information—such as large language models (LLMs) or domain-specific textual knowledge—to guide more precise vision-language multiple instance learning (VLMIL) \cite{shi2024vila}. In the pathology domain, such methods have been increasingly proposed and have demonstrated promising results across various diagnostic tasks by leveraging LLMs.

Despite recent progress in integrating multimodal guidance through LLMs, the scarcity of annotated pathological data remains a fundamental challenge. In this context, few-shot learning has emerged as a promising approach to enhance model generalization under limited supervision. By incorporating strategies such as prototype learning, meta-learning, or transfer learning, few-shot learning can uncover latent feature patterns and improve the model's ability to recognize novel categories, thereby alleviating the performance bottleneck caused by annotation constraints \cite{qu2024pathology}.


As illustrated in Fig.~\ref{intro}(a-c), previous approaches have not integrated multimodal prototypes and typically rely on text-guided mechanisms to assist the learning of visual features. While these models have demonstrated strong classification performance across various tasks, several limitations remain. First, instance-level descriptions generated by LLMs often suffer from inaccuracies due to the lack of fine-grained medical knowledge. Manually crafted prompts that use only class names lack pathological prior knowledge and offer limited discriminative guidance, particularly for ambiguous categories. Second, existing VLMIL frameworks predominantly adopt unidirectional guidance (e.g., text-to-image cross-modal attention or similarity learning), which restricts the potential for synergistic, bidirectional interactions \cite{huang2023visual,li2023task}. This limits effective information exchange, weakens collaborative reasoning, and results in shallow modality alignment. Consequently, these models fail to fully exploit the mutual enhancement potential of visual and textual representations. Third, current prototype-based learning methods typically rely solely on image-text similarity, neglecting the construction and integration of a unified, scale-invariant multimodal similarity space. Using query similarity as the final decision, without accounting for the deficiencies in the knowledge and embedding accuracy of the textual branch, results in suboptimal alignment and fusion.

To address the mentioned limitations, we propose Libra-MIL, with the following contributions:
\begin{itemize}
    \item We first introduce a \textbf{Task Specific approach for Textual Priors Generation}. Leveraging LLMs, we produce priors at both the bag level (e.g., tumors often exhibit expansive growth) and the instance level (e.g., nuclear division, inflammatory cell infiltration), tailored to specific tasks. This design is inspired by the diagnostic reasoning of pathologists, who integrate the global and local context. By incorporating these textual prompts with visual information, the model is guided to focus on morphological features that are more aligned with the task-specific diagnostic objectives.
    \item \textbf{Dual-Prototype Multimodal Learner} enhances generalization in few-shot learning tasks through prototype-based learning. The modality-specific similarities are then stereoscopically infused within a unified embedding space under an optimal transport framework. This fusion minimizes the matching cost between similarities, enabling structure-aware alignment and integration, thereby improving cross-modal semantic consistency and interoperability.
    \item \textbf{Libra-MIL} consistently outperforms baselines across various few-shot learning settings on three pathological datasets. It surpasses state-of-the-art (SOTA) approaches by an average margin of 2.43\% across accuracy, F1-score, and AUC. Extensive ablation studies further demonstrate the effectiveness of each proposed module, and the framework also offers prototype-based interpretability to enhance model transparency.
\end{itemize}

\begin{figure*}[t!]
\includegraphics[width=0.94\textwidth]{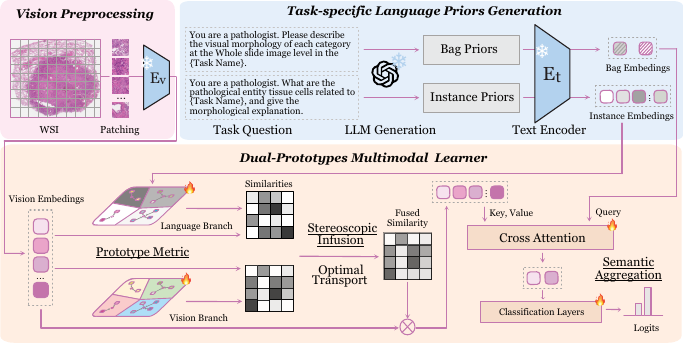} 
\centering
\caption{Overview of Libra-MIL. The framework first employs Vision Preprocessing and Task-specific Language Priors Generation modules to perform modality-specific preprocessing and prior embedding. The dual-prototype multimodal learner then integrates instance-level representations from both modalities into a unified similarity space through prototype-based modeling and Sinkhorn optimal transport.}
\label{methods}
\end{figure*}
\section{Related Work}
\subsection{Vision-Language Multiple Instance Learning}

Vision-Language Multiple Instance Learning (VLMIL) builds upon traditional computational pathology frameworks. Early MIL approaches include ABMIL \cite{ilse2018attention}, which integrated attention mechanisms; CLAM \cite{lu2021data}, enhancing data efficiency through instance clustering and attention pooling; TransMIL \cite{shao2021transmil}, modeling inter-patch dependencies via mutual attention; DSMIL \cite{li2021dual}, fusing multi-scale WSI features with pyramid fusion; H2MIL \cite{hou2022h}, incorporating multi-resolution information through heterogeneous graph learning; and DTFD-MIL \cite{zhang2022dtfd}, employing a pseudo-bag strategy for representation enhancement. Recently, VL frameworks have advanced beyond traditional MIL by integrating vision-language models like CLIP \cite{radford2021learning} and LLMs. TOP-MIL \cite{qu2023rise} leverages CLIP features and GPT-4 linguistic priors for few-shot WSI classification. ViLa-MIL \cite{shi2024vila} fuses vision-language information via prototype-guided decoding and dual-scale text prompts. FOCUS \cite{guo2025focus} combines pathological features with language priors through progressive three-stage compression, while Concept-MIL \cite{sun2025label} predicts pathological concepts to represent key instances through concept-activated patches.

\subsection{Prototypical Networks for Few-shot Learning}
Few-shot learning addresses overfitting in data-scarce scenarios by “learning to learn,” enabling rapid adaptation to novel classes with limited supervision. Prototypical Networks \cite{snell2017prototypical} are a representative approach that learns a metric space where classification is performed based on distances to class prototypes, offering an effective inductive bias for low-data regimes.

Though not explicitly designed for few-shot tasks, several MIL methods have adopted prototype-based mechanisms to improve performance and interpretability. ProtoMIL \cite{rymarczyk2022protomil} incorporates prototype features into instance representations. PAMIL \cite{liu2024pamil} combines prototype learning with attention for bag-level prediction. TP-MIL \cite{yang2023tpmil} optimizes patch-level spaces via instance-prototype distances, while QP-MIL \cite{gou2025queryable} introduces a vision-language queryable prototype framework for incremental WSI classification. These methods highlight prototype learning's broader relevance beyond the few-shot setting.

\subsection{Multimodal Fusion in Computational Pathology}
%

While WSIs are central to computational pathology, the complexity of disease demands multimodal fusion with clinical or genomic data for more comprehensive analysis. MCAT \cite{chen2021multimodal} employs early fusion via co-attention and Set Transformers between image patches and gene embeddings. PathOmics \cite{ding2023pathology} aligns image and omics features in a shared latent space via MSE loss. MOTCat \cite{xu2023multimodal} uses optimal transport-based co-attention for global alignment. SURVPATH \cite{jaume2024modeling} encodes transcriptomic data as pathway-informed tokens and combines them with image patches in a Transformer for survival prediction.


\section{Methodology}
We propose Libra-MIL, a few-shot learning framework for pathological images comprising four key components (Fig.~\ref{methods}): Vision Preprocessing (VP) for WSI feature extraction; Task-specific Language Priors Generation (T-LPG) for generating global and entity-level textual cues; a Dual-Prototype Multimodal Learner (DPM) that aligns visual and textual prototypes via Stereoscopic Optimal Transport (SOT); and Semantic Aggregation (SA) for description-guided querying and information fusion.

\subsection{Preliminary}

MIL is a weakly supervised learning paradigm where data are organized into bags \( B^{(i)} \) for $i^{th}$ sample, each containing instances \( \{\mathbf{x}^{(i)}_{j}\}_{j=1}^{n^{(i)}} \), while only bag-level labels \( y^{(i)} \) (discrete or continuous) are available. The core assumption is that \( y^{(i)} \) depends on the collective properties or the presence of key instances within the bag. This paradigm is suitable for scenarios where instance-level annotations are unavailable. 
\begin{equation}
    f(B^{(i)}) = f(\{x^{(i)}_{1},\cdots,x^{(i)}_{n^{(i)}}\}) \approx y^{(i)}
\end{equation}
$f(\cdot)$ enabling bag-level prediction and implicit instance importance inference. For classification tasks, $y^{(i)} \in \{1,..c\}$.

In \textbf{vision preprocessing}, let a bag $B^{(i)} \in \mathbb{R}^{H\times W \times 3}$ denote a WSI, which is divided into a set of patches $\{ a^{(i)}_{j} \}\in \mathbb{R}^{h\times w\times 3}$ inspired by CLAM \cite{lu2021data}. Due to high computational cost, end-to-end training on pixel-level patches is impractical. A common alternative uses a pretrained vision encoder $E_v$ to extract patch-level features: $\mathbf{x}_{ij} = E_v(a^{(i)}_{j})$.

\subsection{Task-specific Language Priors Generation}
To incorporate high-level semantic cues and pathological domain knowledge, we introduce a T-LPG module that constructs instance-level and bag-level textual descriptions through task-specific prompts.


In the context of a classification task $T$, we design two task-specific prompts based on the \textbf{CHAT} framework \cite{sahoo2024systematic}. One capturing entity-level pathological concepts and the other describing whole-slide morphological patterns. These prompts, denoted by $q \in Q$, are provided as input to a large language model $\text{LLM}(\cdot)$, which simulates a pathologist’s diagnostic reasoning to produce language-based priors. Specifically, the instance-level priors encode fine-grained pathological entities, while the bag-level priors reflect the global morphological context of the slide. The process for generating these two types of priors is:
\begin{equation}
    P_\text{ins}=\text{LLM}(q_\text{ins}), P_\text{bag}=\text{LLM}(q_\text{bag})
\end{equation}
where $P_\text{ins}$ and $P_\text{bag}$ are further embedded into shared semantic space using a frozen text encoder $E_t$:
\begin{equation}
    z_\text{ins} = E_t(P_\text{ins}), z_\text{bag} = E_t(P_\text{bag})
\end{equation}
where $z_\text{ins}\in \mathbb{R}^{k_t\times d}, z_\text{bag} \in \mathbb{R}^{c\times d}$ are the resulting instance-level and bag-level language embeddings. $k_t$ is the number of instance priors. $d$ is the hidden size of $E_t$.
\subsection{Dual-Prototype Multimodal Learner}
To facilitate effective alignment between vision and language modalities under limited supervision in few-shot learning, we propose a two branches in prototype learner that integrates instance-level representations from both branches into a unified similarity space via prototype-based modeling and Stereoscopic Optimal Transport (SOT).


\subsubsection{Prototypes metrics}

To construct a shared semantic space for vision-language alignment, we employ two complementary sets of prototypes: learnable visual prototypes and text prototypes derived from language priors.

Let the set of visual instances bag $B^{(i)}$ be $\mathcal{X}^{(i)} = \{\mathbf{x}_{ij}\}_{j=1}^{n^{(i)}}$, where $\mathbf{x}_{ij} \in \mathbb{R}^{d}$ elements. We introduce a set of learnable vision prototypes $\mathcal{P}_v$ with $K_v$ elements. For multimodal domain-specific prior knowledge, we also introduce text prototypes derived from instance-level textual descriptions. These $K_t$ language priors describe relevant pathological entities' embeddings:
\begin{equation}
\left\{
\begin{aligned}
\mathcal{P}v &= \{ \mathbf{p}_v^{(k_v)} \}_{k_v=1}^{K_v},  &\quad \mathbf{p}_v^{(k_v)} \in \mathbb{R}^d\\
\mathcal{P}t &= \{ \mathbf{p}_t^{(k_t)} = z_{\text{ins}}^{(k_t)} \}_{k_t=1}^{K_t}, &\quad \mathbf{p}_t^{(k_t)} \in \mathbb{R}^d 
\end{aligned}
\right.
\end{equation}
Dual modality prototypes are initialized with $\mathcal{P}_v, \mathcal{P}_t$ and jointly optimized with model parameters. They act as semantic anchors to capture task-relevant instance patterns.

The insance-to-prototype similarity $S^{(i)}(j,k)$ is computed modality-wise as follows:
\begin{equation}
\begin{cases}
\mathbf{S}^{(i)}_{v}(j,k) = \cos(\mathbf{x}_{ij}, \mathbf{p}_v^{(k)}) = \dfrac{\mathbf{x}_{ij}^{\top} \mathbf{p}_v^{(k)}}{\| \mathbf{x}_{ij} \| \cdot \| \mathbf{p}_v^{(k)} \|}, & \mathbf{p}_v^{(k)} \in \mathcal{P}_v \\
\mathbf{S}^{(i)}_{t}(j,k) = \cos(\mathbf{x}_{ij}, \mathbf{p}_t^{(k)}) = \dfrac{\mathbf{x}_{ij}^{\top} \mathbf{p}_t^{(k)}}{\| \mathbf{x}_{ij} \| \cdot \| \mathbf{p}_t^{(k)} \|}, &  \mathbf{p}_t^{(k)} \in \mathcal{P}_t
\end{cases}
\end{equation}
The bimodal similarity quantifies the semantic relevance of each visual instance across different modalities, reflecting its alignment with task-specific concepts. The model is able to enhance  discriminative capability and semantic understanding of individual instances by similarities infusion as follow.
\subsubsection{Stereoscopic Infusion}
Since similarity values are dimensionless and lie in a shared semantic space, we interpret the two similarity matrices—visual-to-visual $\mathbf{S}^{(i)}_{v} \in \mathbb{R}^{n^{(i)}\times K_v}$ and visual-to-textual $\mathbf{S}^{(i)}_{t} \in \mathbb{R}^{n^{(i)}\times K_t}$ as two slices in a conceptual three-dimensional latent space. Each visual instance $\mathbf{x}_{ij}$ can be semantically aligned with prototypes from both visual and textual modalities, forming a multimodal correspondence field.

Inspired by Optimal Transport (OT) \cite{brenier1991polar}, we propose to iteratively fuse $\mathbf{S}_{v}^{(i)}$ and $\mathbf{S}_{t}^{(i)}$ into a unified alignment map by modeling the coupling between instance-level distributions and prototype-level distributions. Specifically, we first define the instance-wise similarity cost matrix $\mathbf{C} \in \mathbb{R}^{K_v \times K_t}$ as the cross-modal discrepancy between vision and text prototypes, computed by:
\begin{equation}
\mathbf{C}(k_v, k_t) = 1 - \cos\left(\mathbf{p}_v^{(k_v)}, \mathbf{p}_t^{(k_t)}\right)
\end{equation}

Then, we define marginal distributions over the two prototype spaces, $\Delta^K$ denoates the K-dimensional probability simplex:
\begin{equation}
\bm{\mu}^{(i)} \in \Delta^{K_v}, \quad \bm{\nu}^{(i)} \in \Delta^{K_t}
\end{equation}
where $\bm{\mu}^{(i)}$ and $\bm{\nu}^{(i)}$ are estimated via average attention distributions from $\mathbf{S}_{v}^{(i)}$ and $\mathbf{S}_{t}^{(i)}$, respectively:
\begin{equation}
\begin{cases}
\bm{\mu}^{(i)} = \mathrm{softmax}\left(\frac{1}{n^{(i)}} \sum_{j=1}^{n^{(i)}} \mathbf{S}_{v}^{(i)}(j,:)\right)\\
\bm{\nu}^{(i)} = \mathrm{softmax}\left(\frac{1}{n^{(i)}} \sum_{j=1}^{n^{(i)}} \mathbf{S}_{t}^{(i)}(j,:)\right)
\end{cases}
\end{equation}

We then solve the entropy-regularized optimal transport problem to find the transport plan $\mathbf{T}^{(i)} \in \mathbb{R}^{K_v \times K_t}$:
\begin{equation}
\mathbf{T}^{(i)} = \mathop{\arg\min}_{\mathbf{T} \in \Pi(\bm{\mu}^{(i)}, \bm{\nu}^{(i)})} \langle \mathbf{T}, \mathbf{C} \rangle - \varepsilon \mathcal{H}(\mathbf{T})
\end{equation}
where $\Pi(\bm{\mu}^{(i)}, \bm{\nu}^{(i)})$ denotes the set of transport plans with marginals $\bm{\mu}^{(i)}$ and $\bm{\nu}^{(i)}$, $\mathcal{H}(\mathbf{T}) = -\sum_{k_v,k_t} \mathbf{T}_{k_v,k_t} \log \mathbf{T}_{k_v,k_t}$ is the entropy regularization term, and $\varepsilon > 0$ controls the smoothness. To efficiently solve this $\mathbf{T}$, we apply the Sinkhorn algorithm \cite{2013Sinkhorn} (Appendix B.1).


Finally, we fused the similarity between instance $\mathbf{x}_{ij}$ and prototype pair $(k_v, k_t)$ is derived by transporting the original similarity maps via the learned plan:
\begin{equation}
\tilde{\mathbf{S}}_{j}^{(i)} = \mathbf{S}_{v}^{(i)}(j,:) \mathbf{T}^{(i)} \mathbf{S}_{t}^{(i)}(j,:)^\top
\end{equation}
This scalar fusion score $\tilde{\mathbf{S}}_{j}^{(i)} \in \mathbb{R}$ acts as a unified multimodal relevance indicator for each instance.
\subsubsection{Semantic Aggregation}
This scalar fusion score is used to reweight the visual instances, and a query-based attention aggregation mechanism, which dynamically summarizes instance information guided by a task-specific query derived from the bag-level textual prior. The fused attention map is obtained by marginalizing over the prompt dimension:
\begin{equation}
\begin{aligned}
\begin{cases}
    \alpha_{\text{fused}}^{(i)} = \sum_j \tilde{\mathbf{S}}_{j,:}^{(i)} 
    \\
    X_{\text{fused}}^{(i)} = \sum_k \alpha_\text{fused}^{(k)} \cdot \mathcal{X}^{(i)} 
\end{cases}
\end{aligned}
\end{equation}
Next, bag priors as query representation $z_\text{bag}$ are used to perform cross-attention over the fused visual tokens:
\begin{equation}
    H^{(i)} = \text{CrossAttn}(z_\text{bag},X_{\text{fused}}^{(i)},X_{\text{fused}}^{(i)})
\end{equation}
Finally, the aggregated bag representation \( H \in \mathbb{R}^{1 \times d} \) is forwarded through a classification head \( g(\cdot) \) followed by an activation function \( \sigma(\cdot) \) to produce the bag-level logits:
\begin{equation}
    \tilde{p}^{(i)}=\sigma(g(H^{(i)}))
\end{equation}
This querying-based aggregation allows the model to focus on task-relevant fine-grained by semantic prompts, enhancing both interpretability and discriminative power.
\subsection{Training Strategy}
Under few-shot weakly supervised learning (FSWL), the model is trained on limited WSIs per class to enhance robustness in low-data scenarios. Libra-MIL is optimized with cross-entropy loss to minimize the negative log-likelihood of ground truth $y^{(i)}$ in a batch $N$:
\vspace{-0.5em}
\begin{equation}
    \mathcal{L}_{CE} = -\frac{1}{N}\sum_{i=1}^{N}y^{(i)}\text{log}(\tilde{p}^{(i)})
\end{equation}
During Stereoscopic Infusion, the Sinkhorn algorithm performs iterative row- and column-wise normalization to satisfy marginal constraints, yielding a differentiable doubly stochastic matrix that minimizes the regularized transport cost for end-to-end training.

\section{Experiments}
\subsection{Settings}
\subsubsection{Datasets.}
We conduct experiments on three publicly available histopathology datasets: \textbf{TCGA-RCC(n=925)}, \textbf{TCGA-NSCLC(n=1052)} \cite{cancergenome2013pan}, and \textbf{CAMELYON16(n=399)} \cite{bejnordi2017diagnostic} . TCGA-RCC contains WSIs of renal cell carcinoma, covering major subtypes of clear cell (KIRC), papillary (KIRP), and chromophobe (KICH). TCGA-NSCLC comprises lung cancer WSIs, including lung adenocarcinoma (LUAD) and lung squamous cell carcinoma (LUSC), providing subtype annotations and high-resolution WSIs suitable for classification. Both datasets are from The Cancer Genome Atlas. CAMELYON16 is a benchmark dataset for detecting breast cancer, consisting of H\&E-stained WSIs with detailed pixel-level annotations of tumor regions. Refer to C.4 for more detailed descriptions.
\subsubsection{Metrics.}
We evaluate all models under 1-, 4-, and 16-shot settings using accuracy (ACC), area under the curve (AUC), and F1-score, averaged over five-fold cross-validation with mean and standard deviation reported. To ensure fair comparison, all methods use identical dataset splits. Results in the table are reported as percentages. Implementation details, environments, and hyperparameters are provided in Appendix C.1, C.2, and C.5.

\begin{table*}[ht!]
\centering 
\small 
\renewcommand{\arraystretch}{1.2} 
\setlength{\tabcolsep}{2.2pt} 
\begin{tabular}{c|cccccccccc}
\toprule 
\multirow{2}{*}{\rotatebox{90}{\textbf{Dataset}}} & \multirow{2}{*}{\textbf{Methods}} & \multicolumn{3}{c}{\textbf{1-shot}} & \multicolumn{3}{c}{\textbf{4-shot}} & \multicolumn{3}{c}{\textbf{16-shot}} \\ \cmidrule(lr){3-5} \cmidrule(lr){6-8} \cmidrule(lr){9-11}
 & & ACC & AUC & F1 & ACC & AUC & F1 & ACC & AUC & F1 \\
\midrule 
\multirow{11}{*}{\rotatebox{90}{TCGA-RCC}}
& Max pooling & 46.7\scriptsize{$\pm$13.9} & 60.6\scriptsize{$\pm$12.7} & 35.1\scriptsize{$\pm$12.4} & 71.4\scriptsize{$\pm$8.8} & 87.2\scriptsize{$\pm$8.2} & 65.9\scriptsize{$\pm$10.1} & 92.8\scriptsize{$\pm$0.9} & 98.2\scriptsize{$\pm$0.3} & 91.0\scriptsize{$\pm$1.4} \\
& ABMIL \cite{ilse2018attention} & 53.7\scriptsize{$\pm$8.1} & 76.5\scriptsize{$\pm$5.6} & 46.8\scriptsize{$\pm$8.1} & 88.4\scriptsize{$\pm$2.5} & 96.8\scriptsize{$\pm$1.3} & 85.2\scriptsize{$\pm$2.6} & 92.8\scriptsize{$\pm$0.8} & 98.4\scriptsize{$\pm$0.6} & 91.0\scriptsize{$\pm$1.6} \\
& TransMIL \cite{shao2021transmil} & 52.8\scriptsize{$\pm$10.2} & 67.8\scriptsize{$\pm$7.0} & 41.3\scriptsize{$\pm$6.6} & 81.2\scriptsize{$\pm$4.5} & 94.2\scriptsize{$\pm$2.9} & 77.4\scriptsize{$\pm$5.3} & 89.5\scriptsize{$\pm$2.0} & 97.8\scriptsize{$\pm$0.7} & 87.2\scriptsize{$\pm$1.8} \\
& CLAM\_SB \cite{lu2021data} & 55.8\scriptsize{$\pm$9.5} & 77.7\scriptsize{$\pm$7.8} & 47.2\scriptsize{$\pm$7.2} & 88.2\scriptsize{$\pm$3.1} & 97.3\scriptsize{$\pm$1.3} & 85.0\scriptsize{$\pm$3.8} & 93.4\scriptsize{$\pm$0.7} & 98.6\scriptsize{$\pm$0.5} & 91.4\scriptsize{$\pm$0.7} \\
& CLAM\_MB \cite{lu2021data} & 56.0\scriptsize{$\pm$9.3} & 79.3\scriptsize{$\pm$7.7} & 46.9\scriptsize{$\pm$10.6} & 89.0\scriptsize{$\pm$2.9} & 96.9\scriptsize{$\pm$1.6} & 85.8\scriptsize{$\pm$3.2} & 93.4\scriptsize{$\pm$1.2} & 98.6\scriptsize{$\pm$0.7} & 91.5\scriptsize{$\pm$1.0} \\
& DSMIL \cite{li2021dual} & 43.7\scriptsize{$\pm$23.6} & 67.6\scriptsize{$\pm$12.6} & 34.2\scriptsize{$\pm$17.9} & 82.0\scriptsize{$\pm$7.6} & 94.8\scriptsize{$\pm$2.2} & 78.0\scriptsize{$\pm$6.8} & 91.5\scriptsize{$\pm$1.3} & 98.2\scriptsize{$\pm$0.7} & 88.9\scriptsize{$\pm$1.9} \\
& TOP-MIL \cite{qu2023rise} & 57.3\scriptsize{$\pm$18.7} & 70.6\scriptsize{$\pm$11.4} & 44.5\scriptsize{$\pm$10.0} & 83.5\scriptsize{$\pm$4.3} & 93.5\scriptsize{$\pm$5.3} & 75.3\scriptsize{$\pm$6.3} & 92.0\scriptsize{$\pm$1.4} & 98.4\scriptsize{$\pm$0.5} & 91.3\scriptsize{$\pm$2.0} \\ 
& PAMIL \cite{liu2024pamil} & 60.1\scriptsize{$\pm$18.2} & 77.9\scriptsize{$\pm$11.5} & \underline{54.4\scriptsize{$\pm$16.4}} & 88.8\scriptsize{$\pm$3.6} & 97.1\scriptsize{$\pm$2.2} & 85.7\scriptsize{$\pm$4.5} & \underline{93.6\scriptsize{$\pm$0.9}} & \underline{98.6\scriptsize{$\pm$0.6}} & \underline{92.0\scriptsize{$\pm$0.7}} \\
& FOCUS \cite{guo2025focus} & \underline{63.2\scriptsize{$\pm$15.5}} & \underline{85.0\scriptsize{$\pm$5.4}} & 54.4\scriptsize{$\pm$15.9} & \underline{89.7\scriptsize{$\pm$3.2}} & \underline{97.6\scriptsize{$\pm$0.9}} & \underline{87.3\scriptsize{$\pm$3.8}} & 92.2\scriptsize{$\pm$0.9} & 98.4\scriptsize{$\pm$0.5} & 90.4\scriptsize{$\pm$1.4} \\ \cmidrule(lr){2-11}
& \textbf{Libra-MIL} & \textbf{71.2\scriptsize{$\pm$12.4}} & \textbf{87.2\scriptsize{$\pm$6.9}} & \textbf{62.4\scriptsize{$\pm$12.6}} & \textbf{91.4\scriptsize{$\pm$3.2}} & \textbf{98.2\scriptsize{$\pm$3.1}} & \textbf{88.5\scriptsize{$\pm$5.3}} & \textbf{93.8\scriptsize{$\pm$0.9}} & \textbf{98.8\scriptsize{$\pm$0.3}} & \textbf{92.3\scriptsize{$\pm$1.4}} \\
\midrule 
\multirow{11}{*}{\rotatebox{90}{NSCLC}}
& Max pooling & 55.5\scriptsize{$\pm$12.8} & 55.4\scriptsize{$\pm$21.6} & 50.8\scriptsize{$\pm$30.8} & 73.5\scriptsize{$\pm$12.2} & 78.1\scriptsize{$\pm$14.5} & 73.5\scriptsize{$\pm$13.4} & 87.7\scriptsize{$\pm$6.1} & 93.5\scriptsize{$\pm$6.1} & 87.8\scriptsize{$\pm$5.6} \\
& ABMIL \cite{ilse2018attention} & 59.4\scriptsize{$\pm$6.8} & 64.6\scriptsize{$\pm$9.9} & 43.7\scriptsize{$\pm$22.0} & 74.5\scriptsize{$\pm$11.6} & 80.9\scriptsize{$\pm$12.4} & 74.3\scriptsize{$\pm$13.5} & 89.6\scriptsize{$\pm$1.9} & 95.6\scriptsize{$\pm$1.6} & 89.5\scriptsize{$\pm$1.7} \\
& TransMIL \cite{shao2021transmil} & 54.5\scriptsize{$\pm$5.3} & 59.0\scriptsize{$\pm$5.7} & 51.0\scriptsize{$\pm$10.5} & 66.1\scriptsize{$\pm$8.2} & 71.6\scriptsize{$\pm$9.7} & 63.2\scriptsize{$\pm$14.4} & 83.4\scriptsize{$\pm$2.5} & 91.1\scriptsize{$\pm$1.8} & 83.1\scriptsize{$\pm$3.0} \\
& CLAM\_SB  \cite{lu2021data} & 55.4\scriptsize{$\pm$4.7} & 65.1\scriptsize{$\pm$7.5} & 48.1\scriptsize{$\pm$19.8} & 76.0\scriptsize{$\pm$8.6} & 83.6\scriptsize{$\pm$8.6} & 76.0\scriptsize{$\pm$8.9} & 89.6\scriptsize{$\pm$2.1} & 96.1\scriptsize{$\pm$2.2} & 89.3\scriptsize{$\pm$1.9} \\
& CLAM\_MB \cite{lu2021data} & 56.4\scriptsize{$\pm$2.6} & \underline{67.7\scriptsize{$\pm$9.4}} & 46.8\scriptsize{$\pm$20.9} & 74.9\scriptsize{$\pm$6.3} & 83.6\scriptsize{$\pm$7.3} & 75.9\scriptsize{$\pm$6.0} & \underline{90.7\scriptsize{$\pm$2.0}} & 96.2\scriptsize{$\pm$2.1} & \textbf{90.4\scriptsize{$\pm$1.9}} \\
& DSMIL \cite{li2021dual} & 55.9\scriptsize{$\pm$11.6} & 60.3\scriptsize{$\pm$13.7} & 34.1\scriptsize{$\pm$29.7} & 67.2\scriptsize{$\pm$16.6} & 71.7\scriptsize{$\pm$19.8} & 62.9\scriptsize{$\pm$20.9} & 86.6\scriptsize{$\pm$5.3} & 92.6\scriptsize{$\pm$4.7} & 86.5\scriptsize{$\pm$5.3} \\
& TOP-MIL \cite{qu2023rise} & \underline{65.4\scriptsize{$\pm$7.1}} & 74.3\scriptsize{$\pm$12.7} & 65.1\scriptsize{$\pm$10.1} & \underline{78.3\scriptsize{$\pm$10.0}} & 84.8\scriptsize{$\pm$10.3} & \underline{78.7\scriptsize{$\pm$9.6}} & 89.0\scriptsize{$\pm$1.8} & 95.9\scriptsize{$\pm$1.3} & 88.6\scriptsize{$\pm$1.8} \\
& PAMIL \cite{liu2024pamil} & 61.2\scriptsize{$\pm$7.9} & 67.0\scriptsize{$\pm$6.1} & \underline{68.9\scriptsize{$\pm$3.0}} & 77.9\scriptsize{$\pm$9.7} & \underline{85.3\scriptsize{$\pm$10.0}} & 78.2\scriptsize{$\pm$10.6} & 89.5\scriptsize{$\pm$2.7} & 96.3\scriptsize{$\pm$2.1} & 89.2\scriptsize{$\pm$2.7} \\
& FOCUS \cite{guo2025focus} & 61.9\scriptsize{$\pm$8.3} & 64.8\scriptsize{$\pm$9.4} & 58.2\scriptsize{$\pm$13.7} & 77.4\scriptsize{$\pm$7.7} & 85.1\scriptsize{$\pm$7.6} & 76.6\scriptsize{$\pm$8.7} & 89.7\scriptsize{$\pm$2.8} & \underline{96.5\scriptsize{$\pm$1.6}} & 89.7\scriptsize{$\pm$2.8} \\ \cmidrule(lr){2-11}
& \textbf{Libra-MIL} & \textbf{69.2\scriptsize{$\pm$9.2}} & \textbf{75.5\scriptsize{$\pm$11.9}} & \textbf{70.6\scriptsize{$\pm$9.1}} & \textbf{83.2\scriptsize{$\pm$6.5}} & \textbf{91.5\scriptsize{$\pm$4.9}} & \textbf{81.1\scriptsize{$\pm$9.7}} & \textbf{90.8\scriptsize{$\pm$1.4}} & \textbf{96.8\scriptsize{$\pm$1.1}} & \textbf{90.4\scriptsize{$\pm$1.5}} \\
\midrule 
\multirow{11}{*}{\rotatebox{90}{CAMELYON16}}
& Max pooling & 49.8\scriptsize{$\pm$11.7} & \underline{57.0\scriptsize{$\pm$8.0}} & 40.5\scriptsize{$\pm$23.3} & \underline{64.7\scriptsize{$\pm$7.7}} & \underline{64.3\scriptsize{$\pm$11.0}} & 50.3\scriptsize{$\pm$11.2} & 88.7\scriptsize{$\pm$2.9} & 91.9\scriptsize{$\pm$4.2} & 84.4\scriptsize{$\pm$4.0} \\
& ABMIL \cite{ilse2018attention} & 53.2\scriptsize{$\pm$9.3} & 46.2\scriptsize{$\pm$11.7} & 21.6\scriptsize{$\pm$14.1} & 52.7\scriptsize{$\pm$9.1} & 47.2\scriptsize{$\pm$11.4} & 19.5\scriptsize{$\pm$11.7} & 88.7\scriptsize{$\pm$3.0} & 89.6\scriptsize{$\pm$4.5} & 83.4\scriptsize{$\pm$4.8} \\
& TransMIL \cite{shao2021transmil} & \underline{61.1\scriptsize{$\pm$2.7}} & 55.6\scriptsize{$\pm$8.1} & 24.4\scriptsize{$\pm$22.6} & 55.8\scriptsize{$\pm$7.7} & 51.6\scriptsize{$\pm$12.1} & 39.9\scriptsize{$\pm$14.4} & 62.2\scriptsize{$\pm$2.5} & 57.0\scriptsize{$\pm$7.3} & 34.7\scriptsize{$\pm$9.0} \\
& CLAM\_SB \cite{lu2021data} & 54.7\scriptsize{$\pm$9.3} & 47.0\scriptsize{$\pm$10.7} & 25.2\scriptsize{$\pm$22.7} & 55.0\scriptsize{$\pm$9.8} & 55.3\scriptsize{$\pm$11.0} & 24.9\scriptsize{$\pm$18.1} & 88.8\scriptsize{$\pm$5.7} & 93.1\scriptsize{$\pm$4.1} & 84.7\scriptsize{$\pm$7.1} \\
& CLAM\_MB \cite{lu2021data} & 48.2\scriptsize{$\pm$7.6} & 43.4\scriptsize{$\pm$5.7} & 36.2\scriptsize{$\pm$17.4} & 52.1\scriptsize{$\pm$10.1} & 50.4\scriptsize{$\pm$10.7} & 19.5\scriptsize{$\pm$12.8} & 89.8\scriptsize{$\pm$4.3} & 93.5\scriptsize{$\pm$3.5} & 85.9\scriptsize{$\pm$5.5} \\
& DSMIL \cite{li2021dual} & 51.0\scriptsize{$\pm$7.7} & 54.3\scriptsize{$\pm$6.0} & 37.5\scriptsize{$\pm$21.5} & 51.8\scriptsize{$\pm$9.6} & 51.7\scriptsize{$\pm$11.9} & 31.9\scriptsize{$\pm$21.8} & 63.7\scriptsize{$\pm$6.6} & 62.5\scriptsize{$\pm$7.8} & 52.9\scriptsize{$\pm$7.8} \\
& TOP-MIL \cite{qu2023rise} & 53.5\scriptsize{$\pm$9.9} & 49.1\scriptsize{$\pm$10.2} & 42.2\scriptsize{$\pm$20.6} & 59.7\scriptsize{$\pm$15.4} & 60.3\scriptsize{$\pm$14.8} & 45.7\scriptsize{$\pm$17.3} & 89.2\scriptsize{$\pm$2.8} & 93.2\scriptsize{$\pm$1.7} & 84.0\scriptsize{$\pm$5.0} \\
& PAMIL \cite{liu2024pamil} & 46.8\scriptsize{$\pm$12.2} & 52.3\scriptsize{$\pm$10.3} & 43.5\scriptsize{$\pm$24.4} & 57.1\scriptsize{$\pm$8.7} & 57.0\scriptsize{$\pm$11.2} & 42.4\scriptsize{$\pm$12.2} & 90.5\scriptsize{$\pm$4.5} & \underline{95.0\scriptsize{$\pm$3.1}} & 87.3\scriptsize{$\pm$5.3} \\
& FOCUS \cite{guo2025focus} & 51.6\scriptsize{$\pm$6.2} & 47.7\scriptsize{$\pm$8.5} & \underline{46.2\scriptsize{$\pm$6.4}} & 61.2\scriptsize{$\pm$14.8} & 61.5\scriptsize{$\pm$15.3} & \underline{56.0\scriptsize{$\pm$15.5}} & \underline{90.9\scriptsize{$\pm$1.0}} & 93.7\scriptsize{$\pm$2.7} & \underline{89.9\scriptsize{$\pm$1.1}} \\ \cmidrule(lr){2-11}
& \textbf{Libra-MIL} & \textbf{62.0\scriptsize{$\pm$5.3}} & \textbf{64.9\scriptsize{$\pm$9.9}} & \textbf{46.6\scriptsize{$\pm$16.5}} & \textbf{65.0\scriptsize{$\pm$8.5}} & \textbf{64.8\scriptsize{$\pm$7.1}} & \textbf{57.2\scriptsize{$\pm$10.3}} & \textbf{91.5\scriptsize{$\pm$4.7}} & \textbf{95.7\scriptsize{$\pm$5.1}} & \textbf{90.6\scriptsize{$\pm$6.2}} \\
\bottomrule 
\end{tabular}
\caption{Comparison of classification performance under 1-, 4-, and 16-shot settings across all baselines. Best results are \textbf{bolded}, second-best are \underline{underlined}. Statistical significance (p$<$0.05) is determined by paired t-tests with 95\% confidence intervals over repeated runs.}
\label{compare}
\end{table*}

\begin{table}[ht!]
\centering 
\small 
\renewcommand{\arraystretch}{1.2} 
\setlength{\tabcolsep}{4pt} 
\begin{adjustbox}{width=1\linewidth}
\begin{tabular}{cccccccccc}
\toprule 
\multirow{2}{*}{\textbf{Methods}} & \multicolumn{3}{c}{\textbf{1-shot}} & \multicolumn{3}{c}{\textbf{4-shot}} & \multicolumn{3}{c}{\textbf{16-shot}} \\ \cmidrule(lr){2-4} \cmidrule(lr){5-7} \cmidrule(lr){8-10}
 & ACC & AUC & F1 & ACC & AUC & F1 & ACC & AUC & F1 \\
\midrule 
w/o instance priors in T-LPG & 67.0 & 83.6 & 56.1 & 87.8 & 96.9 & 84.3 & 92.0 & 98.5 & 90.7 \\
w/o vision prototypes  & 66.0 & 84.7 & 55.5 & 87.2 & 97.4 & 83.7 & 90.3 & 98.0 & 87.5 \\
SOT $\rightarrow$ concat & 66.4 & 81.2 & 54.1 & 86.3 & 97.1 & 82.3 & 92.0 & 98.4 & 90.4 \\
SOT $\rightarrow$ crossAttn & 65.8 & 80.9 & 53.3 & 88.8 & 97.0 & 85.6 & 92.1 & 98.5 & 90.9 \\
w/o bag priors in SA & 55.4 & 82.4 & 47.3 & 88.0 & 97.0 & 88.5 & 92.3 & 98.4 & 90.2 \\ \midrule
\textbf{Libra-MIL} & \textbf{71.2} & \textbf{87.2} & \textbf{62.4} & \textbf{91.4} & \textbf{98.2} & \textbf{88.5} & \textbf{93.8} & \textbf{98.8} & \textbf{92.3} \\
\bottomrule 
\end{tabular}
\end{adjustbox}
\caption{Ablation Studies of Libra-MIL without proposed components on TCGA-RCC.}
\label{Ablation}
\end{table}

\subsection{Comparison Results}
We evaluated Libra-MIL against several competitive methods under K-shot (1, 4, 16) few-shot learning settings, as detailed in Table \ref{compare}. The baseline methods included Max Pooling, ABMIL, TransMIL, CLAM, DSMIL, PAMIL, FOCUS, and TOP-MIL. Across these diverse settings, Libra-MIL demonstrably improved average ACC by 2.49\%, AUC by 2.61\%, and F1-score by 2.15\% compared to the SOTA, and Libra-MIL's performance superiority persists with an increasing number of available shots. We also compare the computational efficiency of methods in Appendix C.7. 

While methods such as DSMIL, FOCUS, TOP-MIL, and Libra-MIL all incorporate multi-scale information, and the latter three all contain text priors. Our approach uniquely integrates multi-modal prototype-based Multiple Instance Learning (MIL). This addresses task-specific text and vision prior bias and often encountered in few-shot scenarios.

\subsection{Ablation and Hyperparameters Studies}
\subsubsection{Modules.}
To further explore the proposed components with the effectiveness of Libra-MIL, we performed the single module ablation experiments in Table \ref{Ablation}. The ablation studies on TCGA-RCC in different few-shot settings investigate the influence of text-priors, vision prototypes, similarities stereoscopic infusion and bag query. The results demonstrate consistent performance improvements with each additional component. In the extremely one-shot setting (w/o Bag Querying), guidance from global information is crucial. The performance drop when replacing optimal transport with concatenation and cross-attention highlights the advantage of the proposed module. (Appendix C.5 for full).

\subsubsection{Pathology Pre-training Models.}
We further investigated the impact of cross-modal latent space misalignment by varying the visual encoder while keeping the text encoder constant. As shown in Table \ref{FM}, different pathology foundation models (Phikon \cite{filiot2023scaling}, GigaPath \cite{xu2024whole}, UNI \cite{chen2024towards}) exhibited significant performance disparities under various few-shot settings compared to CONCH, despite utilizing the same language prior. This highlights the sensitivity of multimodal prototype learning to the alignment of the underlying feature spaces and suggests that cross-modal consistency plays a crucial role in MIL performance. The results underscore the importance of designing robust fusion mechanisms capable of addressing the heterogeneity of cross-modal representations, which is also one of the limitations of our approach.
\begin{table}[ht!]
\centering 
\small 
\renewcommand{\arraystretch}{1.2} 
\setlength{\tabcolsep}{4.5pt} 
\begin{adjustbox}{width=1\linewidth}
\begin{tabular}{cccccccccc}
\toprule 
\multirow{2}{*}{\textbf{Models}} & \multicolumn{3}{c}{\textbf{1-shot}} & \multicolumn{3}{c}{\textbf{4-shot}} & \multicolumn{3}{c}{\textbf{16-shot}} \\ \cmidrule(lr){2-4} \cmidrule(lr){5-7} \cmidrule(lr){8-10}
 & ACC & AUC & F1 & ACC & AUC & F1 & ACC & AUC & F1 \\
\midrule 
Phikon \cite{filiot2023scaling}& 62.2 & 66.6 & 58.9 & 72.2 & 78.8 & 65.1 & 89.0 & 95.9 & 89.0 \\
GigaPath \cite{xu2024whole}  & 65.1 &  72.1 & 66.2 & 71.1 & 77.7 & 63.1 & 87.2 & 94.5 & 87.3 \\
UNI \cite{chen2024towards} & 65.4 & 68.9 & 63.2 & 72.9 & 79.6 & 71.5 & 88.7 & 95.5 & 88.4 \\
CONCH \cite{lu2024avisionlanguage} & 69.2 & 75.5 & 70.6 & 83.2 & 91.5 & 81.1 & 90.8 & 96.8 & 90.4 \\
\bottomrule 
\end{tabular}
\end{adjustbox}
\caption{Variations in pathology foundation models markedly affect outcomes on the TCGA-NSCLC.}
\label{FM}
\end{table}

\subsubsection{Impact of LLM Knowledge.}
\begin{figure}[h!]
\includegraphics[width=0.48\textwidth]{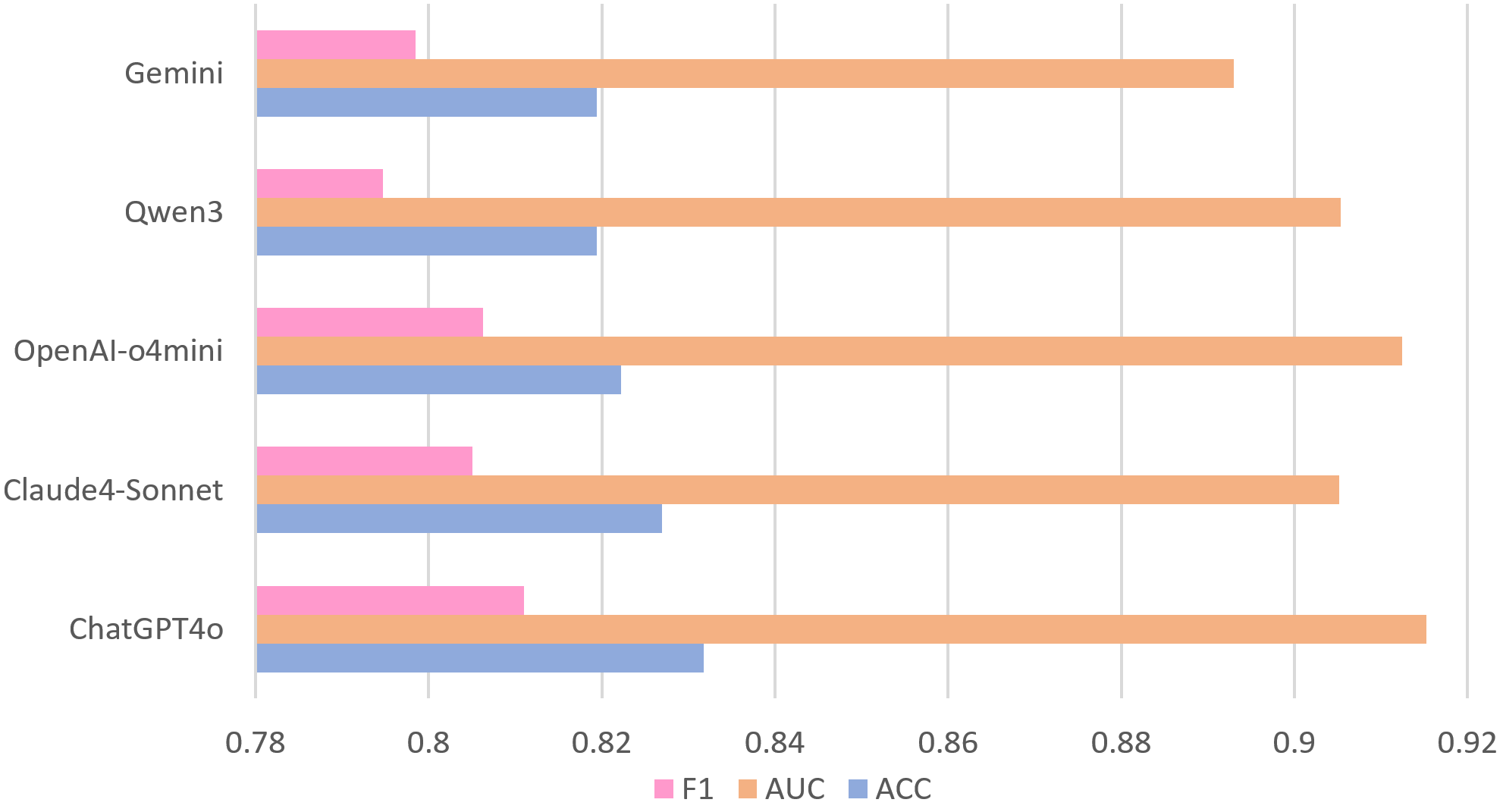} 
\centering
\caption{Results of different LLMs' priors on TCGA-NSCLC under 4-shot setting.}
\label{LLMs}
\end{figure}

To assess how variations in background knowledge across large language models (LLMs) affect task performance, we conduct a comparative study using a unified prompt (see Appendix C.) across several popular LLMs, including Gemini 2.5 \cite{comanici2025gemini}, Qwen3 \cite{yang2025qwen3}, OpenAI-o4mini \cite{menick2024gpt}, Claude4-Sonnet \cite{anthropic2025claude}, and GPT-4o \cite{hurst2024gpt}. Each model generated task-specific textual descriptions at both the instance and bag levels, which are then integrated into the Libra-MIL framework and evaluated under a 4-shot setting on the TCGA-NSCLC dataset. As shown in Fig.~\ref{LLMs}, GPT-4o consistently outperformed other LLMs, while Libra-MIL maintained stable performance across models and consistently surpassed the baseline under identical conditions. These results demonstrate Libra-MIL’s robustness to variations in textual semantics and indicate that more expressive, knowledge-rich LLMs can further enhance performance.

\subsubsection{Vision prototypes Number.}
To assess the influence of the number of vision prototypes on model performance, we conducted ablation studies by varying $K_v$. As shown in the Table \ref{num_prototypes}, the model achieves the best overall performance when $K_v=6$ indicating an optimal trade-off between expressive capacity and generalization. When $K_v$ is small, the model may struggle to capture the diversity of visual patterns. As $K_v$ increases, the performance metrics generally decline, suggesting that introducing too many prototypes may lead to a sparse latent space and ambiguous class boundaries.
\begin{table}[ht]
\centering
\renewcommand{\arraystretch}{1.1}
\setlength{\tabcolsep}{4.5pt}
\begin{adjustbox}{width=1\linewidth}
\begin{tabular}{c|c c c c c c}
\toprule
\textbf{$K_v$ (Vision Prototypes Number)} & 2 & 4 & 6 & 8 & 10 & 12 \\
\midrule
\textbf{ACC}  & 82.7{\scriptsize$\pm$7.7} & 81.8{\scriptsize$\pm$10.2} & 83.2{\scriptsize$\pm$6.5} & 80.0{\scriptsize$\pm$9.8} & 81.6{\scriptsize$\pm$10.3} & 79.5{\scriptsize$\pm$11.6}  \\
\textbf{AUC}  & 90.9{\scriptsize$\pm$6.1} & 89.3{\scriptsize$\pm$9.5} & 91.5{\scriptsize$\pm$4.9} & 88.5{\scriptsize$\pm$9.6} & 89.5{\scriptsize$\pm$9.9} & 87.6{\scriptsize$\pm$11.3} \\
\textbf{F1}  & 81.8{\scriptsize$\pm$10.5} & 79.3{\scriptsize$\pm$15.2} & 81.1{\scriptsize$\pm$9.7} & 80.3{\scriptsize$\pm$9.1} & 82.5{\scriptsize$\pm$8.9} & 76.4{\scriptsize$\pm$18.8}  \\

\bottomrule
\end{tabular}
\end{adjustbox}
\caption{Performance with varying number of visual prototypes under 4-shot setting on TCGA-NSCLC.}
\label{num_prototypes}
\end{table}
\subsection{Multimodal Prototypes with Interpretability.}
\begin{figure}[h!]
\includegraphics[width=0.48\textwidth]{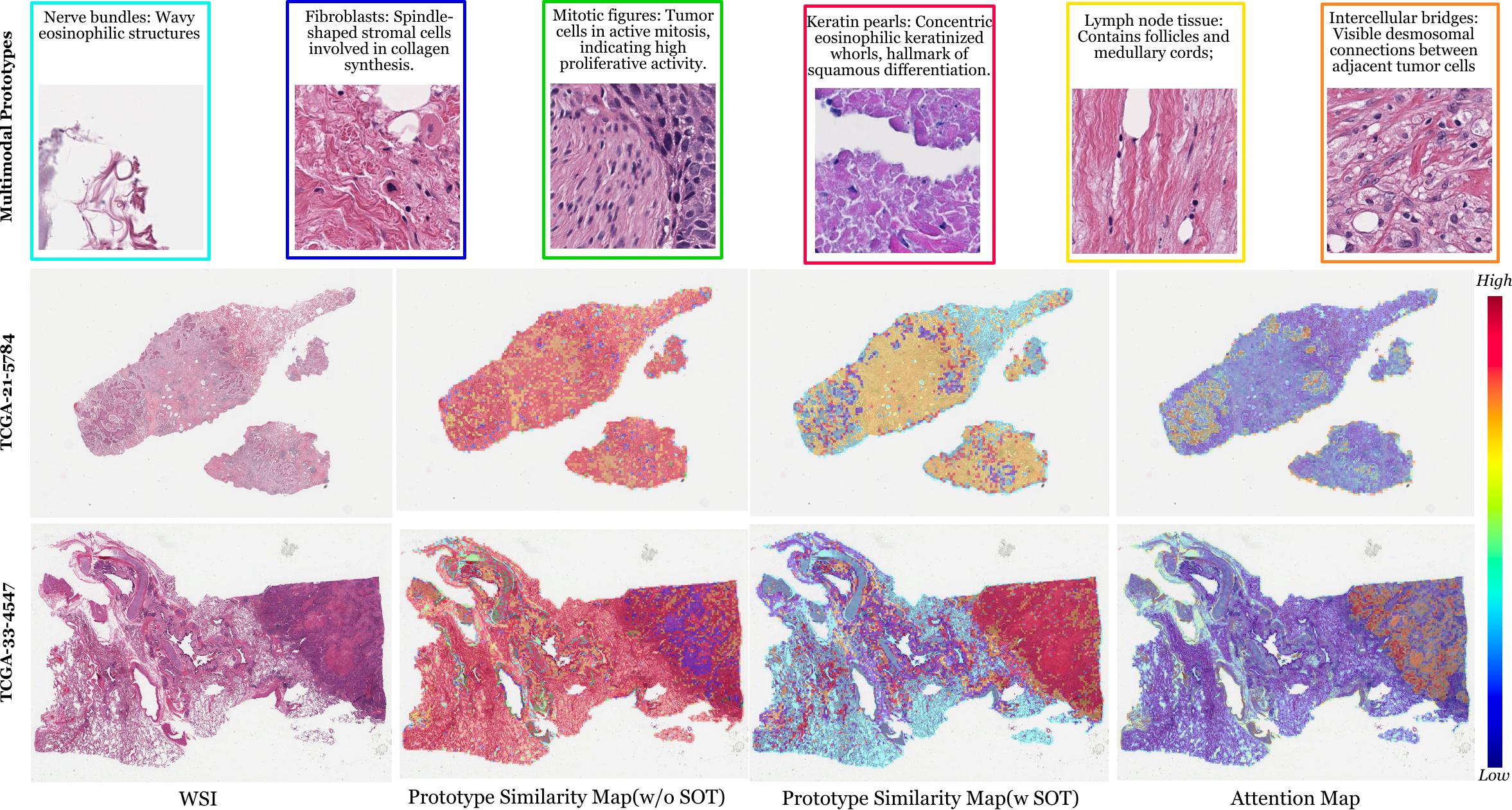} 
\centering
\caption{Case study of multimodal prototypes with different histological morphologies and semantic attention on WSIs.}
\label{proto}
\end{figure}

To llustrate the interpretability of Libra-MIL, we visualize the most representative visual patches as visual prototypes, along with their corresponding textual prototypes, under the $K_v=6$ setting in Fig.~\ref{proto}. These prototypes capture meaningful histological features—such as nerve bundles, mitotic figures, and keratin beads—demonstrating clear diagnostic relevance. Additionally, we present prototype similarity matrices and attention maps of bag-level priors for two TCGA-NSCLC cases, showing that the model consistently attends to clinically significant regions. These results underscore the capability of Libra-MIL’s multimodal prototypes to capture discriminative patterns and enhance interpretability.

\subsection{Joint Contribution of Multimodal Prototypes}
\begin{figure}[h!]
\includegraphics[width=0.48\textwidth]{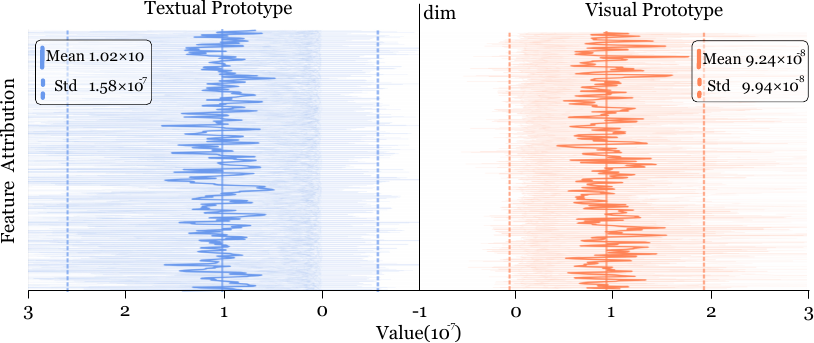} 
\centering
\caption{Gradient-based Contribution of textual and visual prototypes.}
\label{attr}
\end{figure}
To investigate the influence of different modality-specific prototypes on the final diagnostic decision, we applied gradient-based analysis \cite{lundberg2020local2global} of the Libra-MIL (Fig.~\ref{attr}). Specifically, in the 4-shot setting on the TCGA-RCC dataset, we backpropagated the classification output to the prototype embeddings to assess the contribution of each modality-specific prototype across feature dimensions. The results show that the average gradient magnitudes of visual and textual prototypes are comparable, indicating that both modalities play equally important roles in the decision-making process.
\section{Conclusion}
We present Libra-MIL, a multimodal prototype-based MIL framework designed for computational pathology under limited supervision. To overcome key limitations of existing VLMIL methods—such as biased LLM-generated descriptions and unidirectional alignment—we introduce task-specific textual priors and propose a bidirectional fusion strategy based on Stereoscopic Optimal Transport. By constructing both visual and textual prototypes, Libra-MIL enables interpretable and structure-aware cross-modal reasoning. Few-shot experiments on three cancer datasets demonstrate its superior performance and enhanced interpretability. Future work will explore rapid domain-adaptive tuning, broader modality integration, and extensions to more diverse tasks such as survival regression analysis.

\section{Acknowledgments}
This work was supported by National Natural Science Foundation of China (Grant No. 62371409) and Fujian Provincial Natural Science Foundation of China (Grant No. 2023J01005).
\bibliography{aaai2026}
\def\maketitlesupplementary
{
\newpage
   \twocolumn[
    \centering
    \Large
    \vspace{0.5em}\textbf{Supplementary Material} \\
    \vspace{1.0em}
   ] 
}
\clearpage
\setcounter{page}{1}
\maketitlesupplementary
\appendix

Appendix A provides a detailed introduction to the main baseline algorithms for comparison. Theoretical foundations of our approach are further elaborated in Appendix B. Appendix C presents experimental settings, hyperparameters, additional results, and the acceptability analysis. Finally, Appendix D discusses the limitations of our method and potential directions for future work. 

\section{A. Implementation Details of Related Works}
\subsection{A.1 Max Pooling.} In Multiple Instance Learning (MIL), the max pooling operator aggregates instance-level features $\{h_i\}_{i=1}^N$ within a Bag $B$ into a single bag-level representation $z$ as follows:
\begin{equation}
    \mathbf{x} = \max_{i=1,\cdots, N}{\mathbf{h}_i},
\end{equation}
The max operation is applied element-wise across instance features \(\mathbf{h}_i \in \mathbb{R}^d\), yielding the bag-level representation \(\mathbf{x}\). This is then fed into a classifier \(g(\cdot)\) to predict the bag label \(\hat{y} = g(\mathbf{x})\). Max pooling highlights the most salient feature in each dimension, consistent with the MIL assumption that the presence of at least one positive instance determines the bag label.

\subsection{A.2 ABMIL.} In Attention-based Multiple Instance Learning (ABMIL), instance embeddings $\{\mathbf{h}_i\}_{i=1}^{N}$ are aggregated into a bag representation $\mathbf{z}$ via a learned attention mechanism:
\begin{equation}
    a_i = \frac{\exp(\mathbf{w}^\mathrm{T} \tanh({\mathbf{V}\mathbf{h}_i^\mathrm{T}}))}{\sum_{j=1}^N \exp(\mathbf{w}^\mathrm{T} \tanh({\mathbf{V}\mathbf{h}^\mathrm{T}_j}))}
\end{equation}
where $\mathbf{V}$ and $\mathbf{w}$ are learnable parameters. The bag representation is then computed as the weighted sum: $\mathbf{z} = \sum_{i=1}^N a_i \mathbf{h}_i$. This attention mechanism allows the model to softly select and weight the contributions of individual instances according to their relevance for the bag-level task.

\subsection{A.3 TransMIL.} TransMIL applies a Transformer architecture to model relationships between patches in a Whole Slide Image (WSI). The core algorithm begins by creating a location-aware input for each patch. It fuses the patch's features, $\mathbf{h}_k$, with an embedding of its 2D coordinates, $p_k$, to form a final embedding $e_k = \mathbf{h}_k + \mathbf{p}_k$. This sequence of embeddings is then processed by a Transformer Encoder, which uses its self-attention mechanism to globally model how every patch relates to every other, defined by the formula 
\begin{equation}
\text{Attention}(Q, K, V) = \text{softmax}\left(\frac{QK^T}{\sqrt{d_k}}\right)V
\end{equation}
Finally, the resulting context-aware features, $z_k$, are aggregated into a single bag representation, 
\begin{equation}
B_{\text{rep}} = \sum_{k} a_k \mathbf{z}_k
\end{equation}
using a learned attention mechanism. A dual-head classifier leverages this representation for the final bag prediction while also providing instance-level scores, a design which enables deep supervision during training and generates interpretable heatmaps for analysis.

\subsection{A.4 CLAM} formulates whole slide image (WSI) classification as a Multiple Instance Learning (MIL) problem, where each WSI (bag) \( B = \{\mathbf{x}_i\}_{i=1}^N \) contains \(N\) instances \(\mathbf{x}_i\) without instance-level labels, only bag-level label \(Y \in \{0,1\}\). It integrates attention-based instance weighting with sub-bag clustering, enhancing robustness and interpretability in weakly supervised WSI classification.

An attention module computes normalized attention weights \(\alpha_i\) over instances:
\begin{equation}
H = \sum_{i=1}^N \alpha_i \mathbf{h}_i
\end{equation}

The bag \(B\) is partitioned into \(K\) sub-bags \(\{B_k\}_{k=1}^K\), each with representation \(H_k\) computed by the same attention mechanism. A clustering loss encourages sub-bags from the same class to have consistent representations:
\begin{equation}
\mathcal{L}_{cluster} = \frac{1}{K(K-1)} \sum_{\substack{k,k' = 1 \\ k \neq k'}}^K \| H_k - H_{k'} \|_2^2
\end{equation}

A classifier \(g_\phi\) predicts the bag label from \(H\):
\begin{equation}
\hat{Y} = g_\phi(H)
\end{equation}
The overall loss combines cross-entropy and clustering terms:
\begin{equation}
\mathcal{L} = \mathcal{L}_{CE}(\hat{Y}, Y) + \lambda \mathcal{L}_{cluster}
\end{equation}
where \(\lambda > 0\) balances the contribution of the clustering loss.

\textbf{CLAM\_MB} divides the whole bag into multiple independent sub-bags processed separately, while CLAM\_SB partitions the bag into sub-bags with a clustering constraint to enforce consistency and improve feature robustness across sub-bags.

\subsection{A.5 DSMIL} DSMIL simultaneously models instance-level and bag-level information to improve classification under weak supervision. Each instance in a bag \( B = \{\mathbf{x}_i\}_{i=1}^N \) is first encoded into a feature vector by a shared encoder \(f_\theta\):
\begin{equation}
\mathbf{h}_i = f_\theta(\mathbf{x}_i)
\end{equation}

An instance-level scorer \(g_w\) assigns a relevance score \(s_i\) to each feature, which is normalized into attention weights \(\alpha_i\) via softmax:
\begin{equation}
s_i = g_w(\mathbf{h}_i), \quad \alpha_i = \frac{\exp(s_i)}{\sum_{j=1}^N \exp(s_j)}
\end{equation}

The attention weights \(\alpha_i\) reflect the importance of each instance within the bag. DSMIL is trained with a joint loss combining bag-level classification loss \(\mathcal{L}_{\mathrm{bag}}\) and instance-level supervision loss \(\mathcal{L}_{\mathrm{instance}}\), weighted by a hyperparameter \(\lambda\):
\begin{equation}
\mathcal{L} = \mathcal{L}_{\mathrm{bag}}(\hat{Y}, Y) + \lambda \mathcal{L}_{\mathrm{instance}}
\end{equation}

This dual-stream approach enhances interpretability by identifying key instances and achieves robust bag-level predictions under weak supervision.
\textbf{A.6 TOP-MIL}
\paragraph{Dual-Level Prompt Construction (GPT-4 Driven).}

\textbf{(1) Instance-Level Prompt Set:}  
Each prompt set corresponds to a specific tissue phenotype (e.g., lymphocytes) and consists of:
\begin{equation}
    \begin{aligned}
    \mathcal{T}_{\text{inst}} = \underbrace{\text{"an image patch of [Lymphocytes]"}}_{\text{phenotype description}} \\
    + \underbrace{\text{GPT-4 visual description}}_{\text{prior knowledge}} + \underbrace{[\mathbf{v}_1] \cdots [\mathbf{v}_M]}_{\text{learnable vectors}} 
    \end{aligned}
\end{equation}
These prompts are embedded by the text encoder \( f_{\text{text}} \) to generate instance-level prototypes:
\begin{equation}
\mathbf{P} = f_{\text{text}}(\mathcal{T}_{\text{inst}}) \in \mathbb{R}^{n_p \times m}
\end{equation}
where \( n_p \) denotes the number of prototypes and \( m \) is the feature dimension.

\textbf{(2) Bag-Level Prompt Set:}  
Each prompt encodes the global context of the downstream task:
\begin{equation}
\begin{aligned}
\mathcal{T}_{\text{bag}} = \underbrace{\text{"a WSI of [Lung adenocarcinoma]"}}_{\text{task label}}  \\ + \underbrace{\text{GPT-4 pathology description}}_{\text{prior knowledge}} + \underbrace{[\mathbf{u}_1] \cdots [\mathbf{u}_M]}_{\text{learnable vectors}}
\end{aligned}
\end{equation}
The resulting embedding is:
\begin{equation}
\mathbf{B}_i = f_{\text{text}}(\mathcal{T}_{\text{bag}})
\end{equation}

\paragraph{Prompt-Guided Pooling.}

\textbf{Instance-Prototype Similarity:}  
The similarity matrix between instance features \( \mathbf{Z}_i \in \mathbb{R}^{n_i \times m} \) and instance prototypes \( \mathbf{P} \in \mathbb{R}^{n_p \times m} \) is computed as:
\begin{equation}
\mathbf{W}_i = \text{Softmax}_{\text{column}} \left( \mathbf{Z}_i \mathbf{P}^\top \right) \in \mathbb{R}^{n_i \times n_p}
\end{equation}
\subsection{A.7 PAMIL}
\textbf{Prototype Initialization}
Non-random initialization via hierarchical clustering:
\begin{align*}
&\text{Stage 1: } \mathcal{C}_k = \text{K-means}(V_{\text{slide}}, 10) \quad \forall \text{ slides} \\
&\text{Stage 2: } P = \{p_j\}_{j=1}^m = \text{K-means}\left( \bigcup_k \mathcal{C}_k, m \right)
\end{align*}
where $V_{\text{slide}}$ denotes instance features per slide and $m$ is the prototype count.\\
\textbf{Similarity Matrix Computation}
Cross-attention between instances and prototypes:
\begin{equation}
S = \tanh\left(W_q g(V)\right)^{\top} \odot \sigma\left(W_k g(P)\right)
\end{equation}
\begin{itemize}
    \item $g(\cdot)$: Dimension reduction layer (FCN)
    \item $W_q, W_k \in \mathbb{R}^{d \times d}$: Learnable weights
    \item $\odot$: Element-wise multiplication
    \item $S \in \mathbb{R}^{n \times m}$: Similarity matrix
\end{itemize} 
\textbf{Dual-Branch Attention}
Instance Representation Branch is
\begin{align}
A &= W_c S \quad (W_c \in \mathbb{R}^{m \times c}) \\
Y_{\text{inst}} &= f_{\text{cls1}}\left(\texttt{softmax}(A) \cdot g(V)\right)
\end{align}
\textbf{Prototype Representation Branch:}
\begin{align}
\hat{S} &= \texttt{max-pool}(S) \\
Y_{\text{proto}} &= f_{\text{cls2}}\left(\hat{S} \cdot g(P)\right)
\end{align}
where $c$ is the number of classes.

\subsection{A.8 FOCUS} 
FOCUS leverages large language models (LLMs) to generate pathology-specific textual priors that guide the visual feature extraction. Given a set of pathology-related prompts, the LLM produces descriptive texts reflecting disease characteristics, morphological patterns, and histological subtypes. These textual prompts are encoded using a text encoder \(E_{\mathcal{T}}\) to obtain semantic embeddings \(\mathbf{T}^\mathbf{P}\), which are then concatenated with learnable prompt tokens \(\mathbf{T}^\mathbf{L}\) to form the final textual representation:
\begin{equation}    
\mathbf{T} = [\mathbf{T}^\mathbf{L}; \mathbf{T}^\mathbf{P}] \in \mathbb{R}^{(t_1 + t_2) \times d}
\end{equation}

where \(t_1\) and \(t_2\) denote the lengths of pathology-generated and learnable prompts, respectively, and \(d\) is the embedding dimension.

FOCUS employs a three-stage compression approach to reduce redundant visual tokens while preserving critical pathological information:

\paragraph{(a) Global Redundancy Removal}  
Patch similarity within a sliding window of size \(w=32\) is computed using normalized cosine similarity:
\begin{equation} 
\hat{\mathbf{b}}_i = \frac{\mathbf{b}_i}{\|\mathbf{b}_i\|_2}, \quad S_{ij} = \hat{\mathbf{b}}_i \cdot \hat{\mathbf{b}}_j
\end{equation}
A dynamic threshold \(\tau_g = \mu(S) + \sigma(S)\) is used to identify and remove redundant patches:
\begin{equation}    
R_i = \frac{1}{w} \sum_{j=1}^{w} S_{ij} > \tau_g \implies \text{remove patch } i
\end{equation}

\paragraph{(b) Language-Guided Token Prioritization}  
Cross-modal attention scores between textual tokens \(\mathbf{T}\) and patch embeddings \(\mathbf{B}\) are calculated as:
\begin{equation}    
\mathbf{A} = \text{softmax}\left( \frac{(\mathbf{T}W_q)(\mathbf{B}W_k)^\top}{\sqrt{d}} \right), \quad r_i = \frac{1}{t_1+t_2} \sum_{j=1}^{t_1+t_2} \mathbf{A}_{ji}
\end{equation}
The top-\(k\) tokens are selected according to their relevance scores \(r_i\), where \(k = \min(M_{\max}, \gamma N')\) with \(\gamma=0.8\).

\section{B. Method Supplementary}
\subsection{B.1  Sinkhorn algorithm of Optimal Transportation}
\subsection{Sinkhorn Algorithm for Entropy-Regularized Optimal Transport}

Given two discrete probability distributions $\bm{\mu} \in \Delta^{K_v}$ and $\bm{\nu} \in \Delta^{K_t}$, and a cost matrix $\mathbf{C} \in \mathbb{R}^{K_v \times K_t}$, the entropy-regularized optimal transport (OT) problem is defined as:
\begin{equation}
\mathbf{T}^* = \arg\min_{\mathbf{T} \in \Pi(\bm{\mu}, \bm{\nu})} \langle \mathbf{T}, \mathbf{C} \rangle - \varepsilon \mathcal{H}(\mathbf{T})
\end{equation}
where $\Pi(\bm{\mu}, \bm{\nu})$ is the set of transport plans satisfying marginal constraints, and $\mathcal{H}(\mathbf{T}) = -\sum_{i,j} \mathbf{T}_{i,j} \log \mathbf{T}_{i,j}$ is the entropy term.

\paragraph{Sinkhorn Algorithm.} Let $\mathbf{K} = \exp\left(-\mathbf{C}/\varepsilon\right)$ be the Gibbs kernel. We iteratively update scaling vectors $\mathbf{u}$ and $\mathbf{v}$:
\begin{align}
\mathbf{u}^{(\ell+1)} &= \bm{\mu} \oslash (\mathbf{K} \mathbf{v}^{(\ell)}) \\
\mathbf{v}^{(\ell+1)} &= \bm{\nu} \oslash (\mathbf{K}^\top \mathbf{u}^{(\ell+1)})
\end{align}
where $\oslash$ denotes element-wise division. After $L$ iterations, the transport plan is recovered as:
\begin{equation}
\mathbf{T}^{(L)} = \mathrm{diag}(\mathbf{u}^{(L)}) \cdot \mathbf{K} \cdot \mathrm{diag}(\mathbf{v}^{(L)})
\end{equation}

\begin{algorithm}
\caption{Sinkhorn Algorithm}
\begin{algorithmic}[1]
\STATE \textbf{Input:} Cost matrix $\mathbf{C}$, marginals $\bm{\mu}, \bm{\nu}$, regularization $\varepsilon$, iterations $L$
\STATE \textbf{Output:} Transport plan $\mathbf{T}$
\STATE $\mathbf{K} \gets \exp(-\mathbf{C}/\varepsilon)$
\STATE Initialize $\mathbf{u} \gets \mathbf{1}_{K_v}, \quad \mathbf{v} \gets \mathbf{1}_{K_t}$ \\
\FOR{$\ell = 1$ to $L$}
    \STATE $\mathbf{u} \gets \bm{\mu} \oslash (\mathbf{K} \mathbf{v})$
    \STATE $\mathbf{v} \gets \bm{\nu} \oslash (\mathbf{K}^\top \mathbf{u})$
\ENDFOR
\STATE $\mathbf{T} \gets \mathrm{diag}(\mathbf{u}) \cdot \mathbf{K} \cdot \mathrm{diag}(\mathbf{v})$
\end{algorithmic}
\end{algorithm}

\paragraph{Convergence Guarantee.} If $\mathbf{K} = \exp(-\mathbf{C}/\varepsilon)$ is strictly positive and $\bm{\mu}, \bm{\nu} \in \Delta^K$ have strictly positive entries, then:

\begin{itemize}
  \item The algorithm converges to a unique solution $\mathbf{T}^*$ satisfying:
  \begin{equation}
  \mathbf{T}^* \mathbb{1}_{K_t} = \bm{\mu}, \quad (\mathbf{T}^*)^\top \mathbb{1}_{K_v} = \bm{\nu}
  \end{equation}
  \item The convergence rate is geometric with respect to the Hilbert projective metric.
\end{itemize}

\paragraph{Proof Sketch.} The entropy regularization makes the objective strictly convex. The dual of the regularized OT becomes smooth and unconstrained:
\begin{equation}
\max_{f, g} -\varepsilon \sum_{i,j} \exp\left(\frac{f_i + g_j - C_{i,j}}{\varepsilon}\right) + \langle f, \bm{\mu} \rangle + \langle g, \bm{\nu} \rangle
\end{equation}
Sinkhorn scaling performs coordinate ascent in this dual space. Due to the strict positivity of $\mathbf{K}$, the iterative updates form a contraction mapping in projective space, guaranteeing convergence.

For more detailed theoretical analysis, refer to \cite{2013Sinkhorn, peyre2019computational}.
\subsection{B.2 Stereoscopic Infusion Pseudocodes}
\begin{algorithm}[H]
\caption{Stereoscopic Infusion: Multimodal Similarity Fusion}
\label{alg:stereoscopic_infusion}
\begin{algorithmic}[1]
\REQUIRE Visual similarity matrix $\mathbf{S}_{v}^{(i)} \in \mathbb{R}^{n_i \times K_v}$, textual similarity matrix $\mathbf{S}_{t}^{(i)} \in \mathbb{R}^{n_i \times K_t}$, visual prototypes $\{\mathbf{p}_v^{(k_v)}\}_{k_v=1}^{K_v}$, textual prototypes $\{\mathbf{p}_t^{(k_t)}\}_{k_t=1}^{K_t}$, transport plan $\mathbf{T}^{(i)} \in \mathbb{R}^{K_v \times K_t}$
\ENSURE Fused similarity scores $\{\tilde{\mathbf{S}}_{j}^{(i)}\}_{j=1}^{n_i}$ for instances

\STATE Compute cost $\mathbf{C}(k_v, k_t) = 1 - \cos(\mathbf{p}_v^{(k_v)}, \mathbf{p}_t^{(k_t)})$

\STATE Estimate marginal distributions $\bm{\mu}^{(i)}$ and $\bm{\nu}^{(i)}$ from $\mathbf{S}_{v}^{(i)}$ and $\mathbf{S}_{t}^{(i)}$

\STATE Compute transport plan $\mathbf{T}^{(i)}$ by solving entropy-regularized optimal transport

\FOR{$j = 1$ to $n_i$}
    \STATE Compute fused similarity score:
    \[
    \tilde{\mathbf{S}}_{j}^{(i)} \gets \mathbf{S}_{v}^{(i)}(j,:) \cdot \mathbf{T}^{(i)} \cdot \mathbf{S}_{t}^{(i)}(j,:)^\top
    \]
\ENDFOR

\STATE \RETURN $\{\tilde{\mathbf{S}}_{j}^{(i)}\}_{j=1}^{n_i}$
\end{algorithmic}
\end{algorithm}
\begin{figure*}[ht!]
\includegraphics[width=1\textwidth]{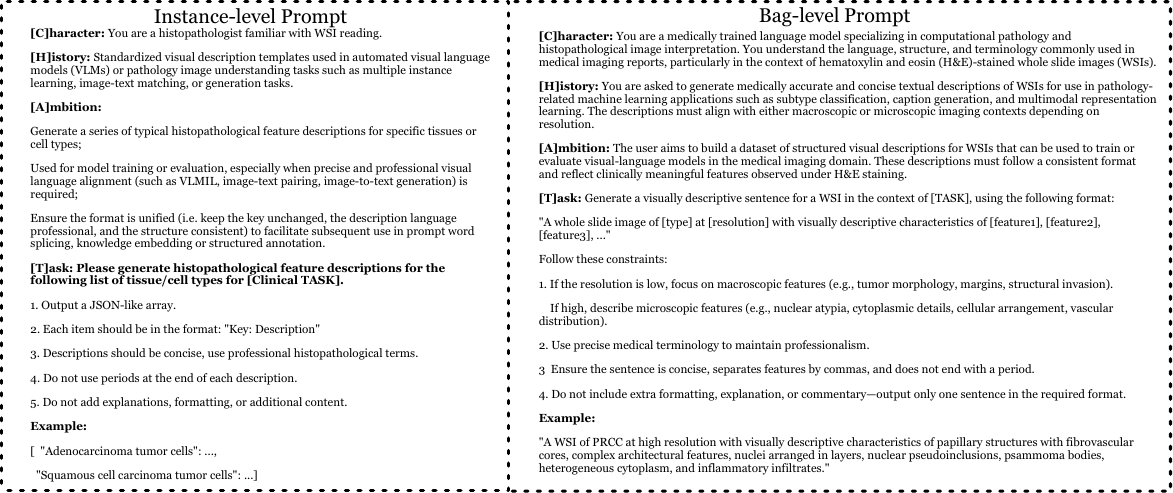} 
\centering
\caption{The prompts of instance and bag priors for the task.}
\label{prompts}
\end{figure*}

\begin{table*}[ht!]
\centering 
\small 
\renewcommand{\arraystretch}{1.2} 
\setlength{\tabcolsep}{4.5pt} 
\begin{tabular}{c|cccccccccc}
\toprule 
\multirow{2}{*}{\rotatebox{90}{\textbf{Dataset}}} & \multirow{2}{*}{\textbf{Methods}} & \multicolumn{3}{c}{\textbf{1-shot}} & \multicolumn{3}{c}{\textbf{4-shot}} & \multicolumn{3}{c}{\textbf{16-shot}} \\ \cmidrule(lr){3-5} \cmidrule(lr){6-8} \cmidrule(lr){9-11}
 & & ACC & AUC & F1 & ACC & AUC & F1 & ACC & AUC & F1 \\
\midrule 
\multirow{6}{*}{\rotatebox{90}{TCGA-RCC}}
& w/o T-LPG & 67.0\scriptsize{$\pm$5.7} & 83.6\scriptsize{$\pm$6.2} & 56.1\scriptsize{$\pm$4.5} & 87.8\scriptsize{$\pm$3.9} & 96.9\scriptsize{$\pm$1.4} & 84.3\scriptsize{$\pm$5.6} & 92.0\scriptsize{$\pm$1.9} & 98.5\scriptsize{$\pm$0.2} & 90.7\scriptsize{$\pm$1.5} \\
& w/o Vision Prototypes  & 66.0\scriptsize{$\pm$2.3} & 84.7\scriptsize{$\pm$6.3} & 55.5\scriptsize{$\pm$7.0} & 87.2\scriptsize{$\pm$4.9} & 97.4\scriptsize{$\pm$1.2} & 83.7\scriptsize{$\pm$6.8} & 90.3\scriptsize{$\pm$4.3} & 98.0\scriptsize{$\pm$11} & 87.5\scriptsize{$\pm$1.5} \\
& SOT $\rightarrow$ Concat & 66.4\scriptsize{$\pm$6.3} & 81.2\scriptsize{$\pm$5.6} & 54.1\scriptsize{$\pm$6.3} & 86.3\scriptsize{$\pm$5.6} & 97.1\scriptsize{$\pm$1.1} & 82.3\scriptsize{$\pm$7.4} & 92.0\scriptsize{$\pm$1.5} & 98.4\scriptsize{$\pm$0.6} & 90.4\scriptsize{$\pm$1.3} \\
& SOT $\rightarrow$ CrossAttn & 65.8\scriptsize{$\pm$9.3} & 80.9\scriptsize{$\pm$6.4} & 53.3\scriptsize{$\pm$12.1} & 88.8\scriptsize{$\pm$3.3} & 97.0\scriptsize{$\pm$1.4} & 85.6\scriptsize{$\pm$6.4} & 92.1\scriptsize{$\pm$1.6} & 98.5\scriptsize{$\pm$0.4} & 90.9\scriptsize{$\pm$2.2} \\
& w/o SA & 55.4\scriptsize{$\pm$14.9} & 82.4\scriptsize{$\pm$6.9} & 47.3\scriptsize{$\pm$10.7} & 88.0\scriptsize{$\pm$4.6} & 97.0\scriptsize{$\pm$3.1} & 88.5\scriptsize{$\pm$5.3} & 92.3\scriptsize{$\pm$1.5} & 98.4\scriptsize{$\pm$0.7} & 90.2\scriptsize{$\pm$1.9} \\ \cmidrule(lr){2-11}
& \textbf{Libra-MIL} & \textbf{71.2\scriptsize{$\pm$12.4}} & \textbf{87.2\scriptsize{$\pm$6.9}} & \textbf{62.4\scriptsize{$\pm$12.6}} & \textbf{91.4\scriptsize{$\pm$3.2}} & \textbf{98.2\scriptsize{$\pm$3.1}} & \textbf{88.5\scriptsize{$\pm$5.3}} & \textbf{93.8\scriptsize{$\pm$0.9}} & \textbf{98.8\scriptsize{$\pm$0.3}} & \textbf{92.3\scriptsize{$\pm$1.4}} \\
\bottomrule 
\end{tabular}
\caption{Ablation Studies of Libra-MIL without proposed components on TCGA-RCC.}
\label{Ablation-appendix}
\end{table*}
\section{C.Experiments Supplementary}
\subsection{C.1 Implementation Details}
For our slide-level few-shot classification experiments, we first preprocess Whole Slide Images (WSIs) into non-overlapping 512$\times$512 pixel tiles at 20$\times$ magnification using the CLAM framework \cite{lu2021data}. One dataset is partitioned using a 5-fold cross-validation scheme, where M samples are randomly drawn from each training fold for the few-shot setting. For each training fold, all hyperparameters are fixed to assess stability and reproducibility.

In baseline comparison, models leverage CONCH as the vision-text foundation towers for embedding. Bag-level tasks and pathological instance textual descriptions are uniformly extracted using the large language model GPT-4o. The model is trained for a maximum of 80 epochs, Adam optimized with a learning rate of $1\times10^{-4}$, employing an early stopping strategy with a patience of 15 to prevent overfitting. 
The model checkpoint yielding the best performance on the validation set is ultimately selected for final evaluation on the test set.
\subsection{C.2 Experiment Environment and Dependencies}
All experiments are conducted on a workstation with 8 NVIDIA GeForce RTX 3090 (24 GB) GPUs and 4 Intel Xeon Silver 4210R CPUs equipped with 256 GB memory. System is on Ubuntu 18.04 LTS. The implementation was based on Python 3.10 and PyTorch 2.1.2, with CUDA 11.8.
\subsection{C.3 More descriptions about datasets}
\textbf{TCGA-RCC} cohort comprises 925 H\&E-stained diagnostic WSIs from three renal cell carcinoma (RCC) subtypes: KICH (109), KIRC (519), and KIRP (297). All slides were obtained from TCGA and scanned at 20$\times$ or 40$\times$ using Aperio or Hamamatsu scanners. Each patient is associated with one diagnostic slide and corresponding clinical metadata (e.g., survival, stage, grade, molecular subtype). To ensure quality and reduce inter-center bias, only primary tumor slides with adequate quality were retained. This dataset is widely used for subtype classification, survival prediction, and MIL research.

\textbf{TCGA-NSCLC} contains 1,052 H\&E-stained WSIs from TCGA, including 540 lung adenocarcinoma (LUAD) and 512 lung squamous cell carcinoma (LUSC) cases. Slides were scanned at 20$\times$/40$\times$ magnification, with one high-quality diagnostic slide per patient selected. Clinical metadata includes survival, TNM stage, histological grade, smoking history, and key mutations (e.g., EGFR, KRAS, TP53). This cohort supports research in subtype classification, prognosis modeling, and weak supervision.

\textbf{CAMELYON16} is a benchmark dataset for metastasis detection in sentinel lymph nodes, containing 399 WSIs from RUMC and UMCU. The training set (270 slides) includes pixel-level metastasis annotations; the test set (129 slides) has hidden labels for challenge evaluation. The number of WSI tumors and normal tissue is 160:239. All slides were scanned at 40$\times$. It is widely used for tumor detection, segmentation, and patch-based classification in lymph node tissues.

\subsection{C.4 Task-specific Prompts and Priors Examples}
For the specific task of pathological multi-instance learning, we propose a prompt engineering strategy based on the \textbf{CHAT framework} to extract textual priors from large language models, covering both instance-level and bag-level descriptions, as illustrated in Fig.~\ref{prompts}.

The \textbf{CHAT} framework consists of four components:
\begin{itemize}
    \item \textbf{Character}: Provides the language model with contextual information about the user's identity and role, allowing it to tailor its responses accordingly. 
    \item \textbf{History}: Supplies relevant background information and contextual knowledge related to the current query, enabling the model to better understand the user's situation and prior interactions.
    \item \textbf{Ambition}: Describes the user's intended long-term or short-term objectives in interacting with the model, helping guide the generation toward more targeted and goal-driven outputs.
    \item \textbf{Task}: Specifies the concrete action or response the user expects from the model, offering the most direct instruction for completing the desired operation.
\end{itemize}

\subsection{C.5 Complete table of ablation experiments}
We show the complete ablation experiment table (including standard deviation) in Table \ref{Ablation-appendix}.
\subsection{C.6 Hyperparameters of Libra-MIL}
Table~\ref{hyperparameters} summarizes the hyperparameter settings used in our method. Notably, we adopt different numbers of multimodal prototypes across datasets, which is determined by the specific characteristics of each task to enable more comprehensive and efficient few-shot learning.
\begin{table}[ht]
\centering
\renewcommand{\arraystretch}{1.1}
\setlength{\tabcolsep}{4.5pt}
\begin{adjustbox}{width=1\linewidth}
\begin{tabular}{c|c c c}
\toprule
\textbf{Hyperparameters} & \textbf{TCGA-RCC} & \textbf{TCGA-NSCLC} & \textbf{CAMELYON16} \\
\midrule
$K_v$  & 10 & 6 & 10  \\
$K_t$  & 46 & 33 & 29 \\
Attention head  & 8 & 8 & 8 \\
$\epsilon$  & 0.05 & 0.05 & 0.05 \\
OT iteration & 20 & 20 & 20 \\
Hidden\_dim & 512 & 512 & 512 \\
\bottomrule
\end{tabular}
\end{adjustbox}
\caption{Hyperparameter settings of Libra-MIL for different datasets.}
\label{hyperparameters}
\end{table}
\subsection{C.7 Computational cost comparison}
As shown in Table \ref{eff}, we compare the average inference time per batch across various MIL methods. Pooling-based models like Max Pooling  and ABMIL exhibit minimal overhead due to their simplicity. Attention-based variants show slightly increased time, reflecting their more complex instance selection.

Transformer-based and prototype-enhanced models, PAMIL (0.003108s), incur higher costs due to deeper architectures and fine-grained relation modeling. FOCUS shows the highest latency (0.549929 s), likely due to global patch-token interactions and sliding-window processing.

Libra-MIL requires 0.006183 s per batch—higher than early baselines but comparable to recent methods like TransMIL and TOP-MIL (0.007162 s). The additional cost stems from its dual-modality processing and optimal transport, but remains practical, balancing efficiency and representational power.
\begin{table}[ht]
\centering
\renewcommand{\arraystretch}{1.1}
\setlength{\tabcolsep}{4.5pt}
\begin{adjustbox}{width=1\linewidth}
\begin{tabular}{c|c}
\toprule
\textbf{Methods} & \textbf{Average calculation time (second/per batch)} \\
\midrule
Max pooling &  0.000503  \\
ABMIL  & 0.000856  \\
TransMIL  & 0.005811 \\
CLAM\_SB & 0.000950  \\
CLAM\_MB & 0.001707 \\
DSMIL &  0.001417 \\
TOP-MIL & 0.007162 \\
PAMIL & 0.003108 \\
FOCUS & 0.549929 \\
\midrule
\textbf{Libra-MIL} & 0.006183 \\
\bottomrule
\end{tabular}
\end{adjustbox}
\caption{Comparison of computational efficiency.}
\label{eff}
\end{table}
\subsection{C.8 More Case Studies of Prototypes with Interpretability}
\begin{figure*}[ht!]
\includegraphics[width=1\textwidth]{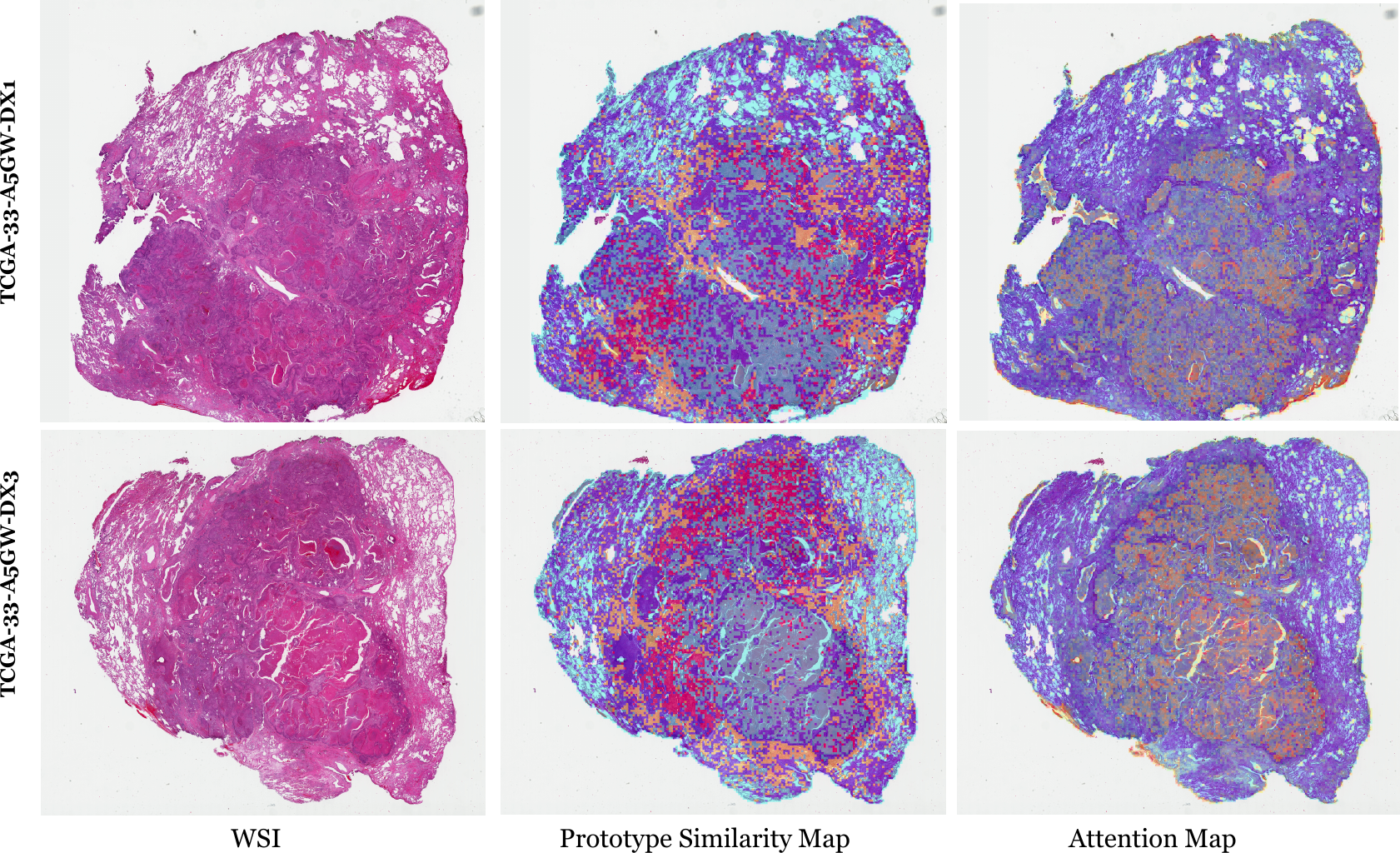} 
\centering
\caption{More case studies of prototypes similarities and task attention.}
\label{appendix-vis}
\end{figure*}

Additional prototype visualization cases and attention maps are in Figure~\ref{appendix-vis}, highlighting how different subtypes within the task attend to distinct instance prototype regions.
\begin{figure*}[ht!]
\includegraphics[width=1\textwidth]{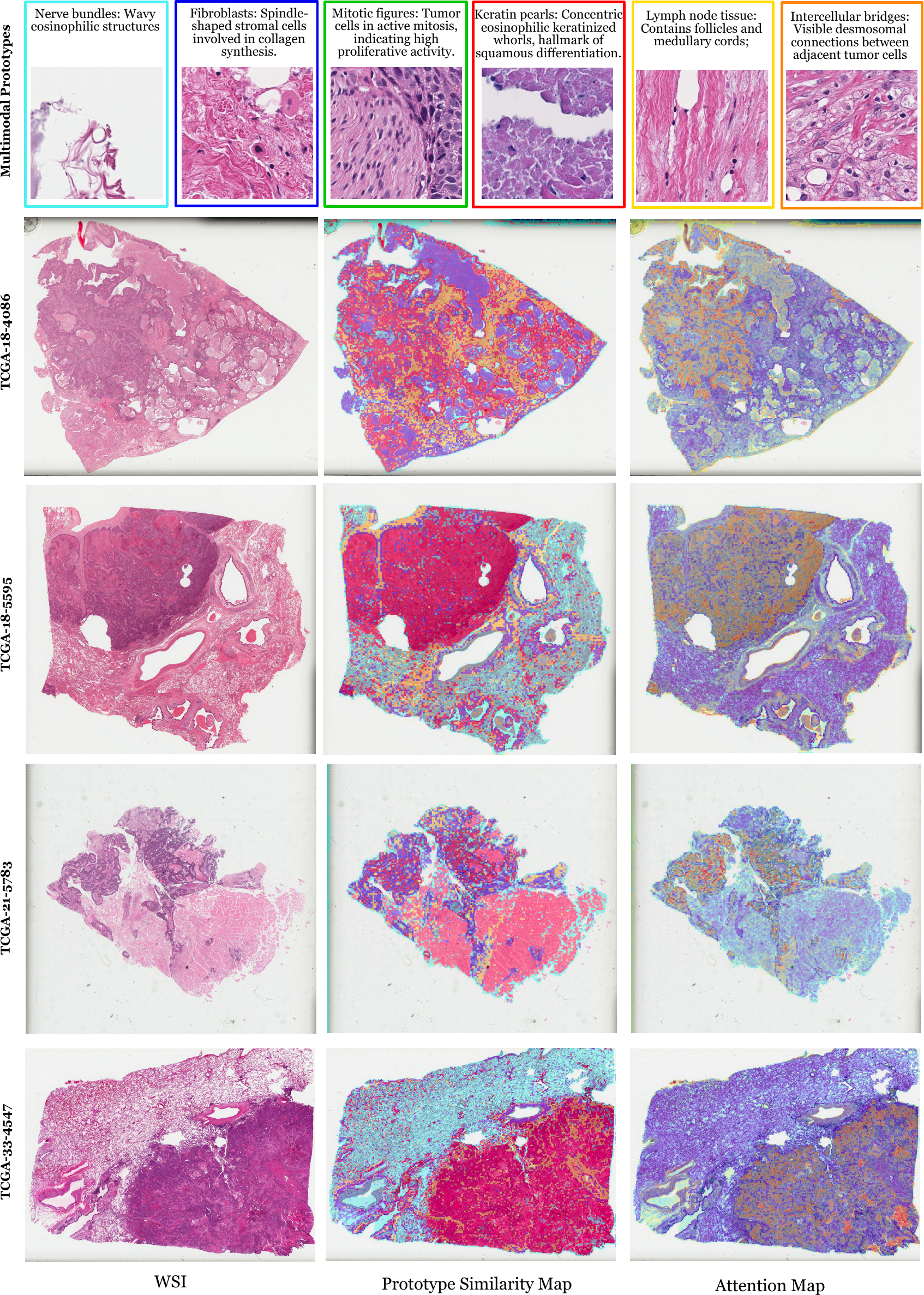} 
\centering
\end{figure*}

\section{D. Limiataions}
\subsection{D.1 Domain-adaptive Learning}
In few-shot multiple instance learning, domain-adaptive learning often faces significant challenges due to the limited availability of target data, which hinders effective distribution alignment—a key factor as evidenced by the results in Table~\ref{FM}. Existing methods typically emphasize global domain shifts while neglecting fine-grained instance-level variability, which is critical in MIL. Moreover, class imbalance, prototype drift, and subtype-specific differences further constrain the adaptability of Libra-MIL.
\subsection{D.2 Lacks task Diversity}
Furthermore, Libra-MIL was only evaluated on classification, which, while essential, covers just a subset of clinical applications. Its performance on other key tasks remains unexplored: a) Segmentation: Identifying and delineating specific regions of interest (e.g., tumor regions, specific tissue types, cellular structures). This is crucial for quantitative analysis, tumor burden assessment, and treatment planning. b) Detection: Localizing specific objects or events within a WSI (e.g., mitotic figures, micro-metastases, specific cell types). This is essential for automated scoring systems and identifying rare events. c)Regression: Predicting continuous values (e.g., tumor cellularity percentage, biomarker expression levels, patient survival time prediction). This provides finer-grained prognostic and predictive information beyond discrete categories. d) Whole Slide Regression/Scoring: Directly predicting a slide-level score (e.g., PD-L1 Combined Positive Score, tumor grade components) without explicit intermediate steps. This is highly relevant in diagnostic workflows.

\end{document}